\documentclass[10pt,twocolumn,letterpaper]{article}

\usepackage[pagenumbers]{cvpr} 

\usepackage[dvipsnames]{xcolor}

\usepackage{graphicx,scalerel}
\usepackage{amsmath}
\usepackage{amssymb}
\usepackage{booktabs}

\newcommand{\hide}[1]{}

\newcommand\wh[1]{\hstretch{2}{\hat{\hstretch{0.5}{#1\mkern1.5mu}}}\mkern0mu}
\usepackage{color}
\usepackage{xcolor,colortbl}
\usepackage{mathtools}
\usepackage{multirow}
\usepackage{bm} 
\definecolor{purple}{rgb}{0.65,0,0.65}
\definecolor{dark_green}{rgb}{0, 0.5, 0}
\definecolor{blueish}{rgb}{0.0, 0.3, .6}
\definecolor{tabhighlight}{HTML}{e5e5e5}
\newcolumntype{h}{>{\columncolor{tabhighlight}}c}

\makeatletter
\newcommand*\wt[2][0.15ex]{%
        \begingroup
        \mathchoice{\wt@helper{#1}{#2}{\displaystyle}{\textfont}}
                   {\wt@helper{#1}{#2}{\textstyle}{\textfont}}
                   {\wt@helper{#1}{#2}{\scriptstyle}{\scriptfont}}
                   {\wt@helper{#1}{#2}{\scriptscriptstyle}{\scriptscriptfont}}%
        \endgroup
        #2%
}
\newcommand*\wt@helper[4]{%
        \def\currentfont{\the#41}%
        \def\currentskewchar{\char\the\skewchar\currentfont}%
        \setbox\tw@\hbox{\currentfont#2\currentskewchar}%
        \dimen@ii\wd\tw@
        \setbox\tw@\hbox{\currentfont#2{}\currentskewchar}%
        \advance\dimen@ii-\wd\tw@
        \rlap{\raisebox{-#1}{$\m@th#3\kern\dimen@ii\widetilde{\phantom{#2}}$}}%
}
\makeatother

\usepackage[export]{adjustbox} 

\newlength{\gridimagewidth}
\usepackage{array}
\newcolumntype{C}[1]{>{\centering\let\newline\\\arraybackslash\hspace{0pt}}m{#1}}

\newcolumntype{h}{>{\columncolor{tabhighlight}}c}

\newcommand{\method}{ARNIQA\xspace}
\newcommand{\methodextended}{ARNIQA (leArning distoRtion maNifold for Image Quality Assessment)\xspace}
\newcommand{\live}{LIVE\xspace}

\newcommand{\csiq}{CSIQ\xspace}
\newcommand{\tid}{TID2013\xspace}
\newcommand{\kadid}{KADID\xspace}

\newcommand{\spaq}{SPAQ\xspace}
\newcommand{\flive}{FLIVE\xspace}
\newcommand{\srcc}{SRCC\xspace}
\newcommand{\plcc}{PLCC\xspace}
\newcommand{\ndist}{n_{dist}\xspace}
\newcommand{\Ndist}{N_{dist}\xspace}

\definecolor{cvprblue}{rgb}{0.21,0.49,0.74}
\usepackage[pagebackref,breaklinks,colorlinks]{hyperref}

\usepackage[capitalize]{cleveref}
\crefname{section}{Sec.}{Secs.}
\Crefname{section}{Section}{Sections}
\Crefname{table}{Table}{Tables}
\crefname{table}{Tab.}{Tabs.}

\def\paperID{*****} 
\def\confName{CVPR}
\def\confYear{2024}

\title{\method: Learning Distortion Manifold for Image Quality Assessment}

\author{Lorenzo Agnolucci \and Leonardo Galteri \and  Marco Bertini \and Alberto Del Bimbo \vspace{0.1ex} \and
University of Florence - Media Integration and Communication Center (MICC) \\
Florence, Italy\\
{\tt\small [name.surname]@unifi.it}
}

\begin{document}
\maketitle
\begin{abstract}
No-Reference Image Quality Assessment (NR-IQA) aims to develop methods to measure image quality in alignment with human perception without the need for a high-quality reference image. In this work, we propose a self-supervised approach named \methodextended for modeling the image distortion manifold to obtain quality representations in an intrinsic manner. First, we introduce an image degradation model that randomly composes ordered sequences of consecutively applied distortions. In this way, we can synthetically degrade images with a large variety of degradation patterns. Second, we propose to train our model by maximizing the similarity between the representations of patches of different images distorted equally, despite varying content. Therefore, images degraded in the same manner correspond to neighboring positions within the distortion manifold. Finally, we map the image representations to the quality scores with a simple linear regressor, thus without fine-tuning the encoder weights. The experiments show that our approach achieves state-of-the-art performance on several datasets. In addition, \method demonstrates improved data efficiency, generalization capabilities, and robustness compared to competing methods. The code and the model are publicly available at \small{\href{https://github.com/miccunifi/ARNIQA}{\url{https://github.com/miccunifi/ARNIQA}}}.
\end{abstract}    
\section{Introduction}
\begin{figure}
    \centering
    \includegraphics[width=\linewidth]{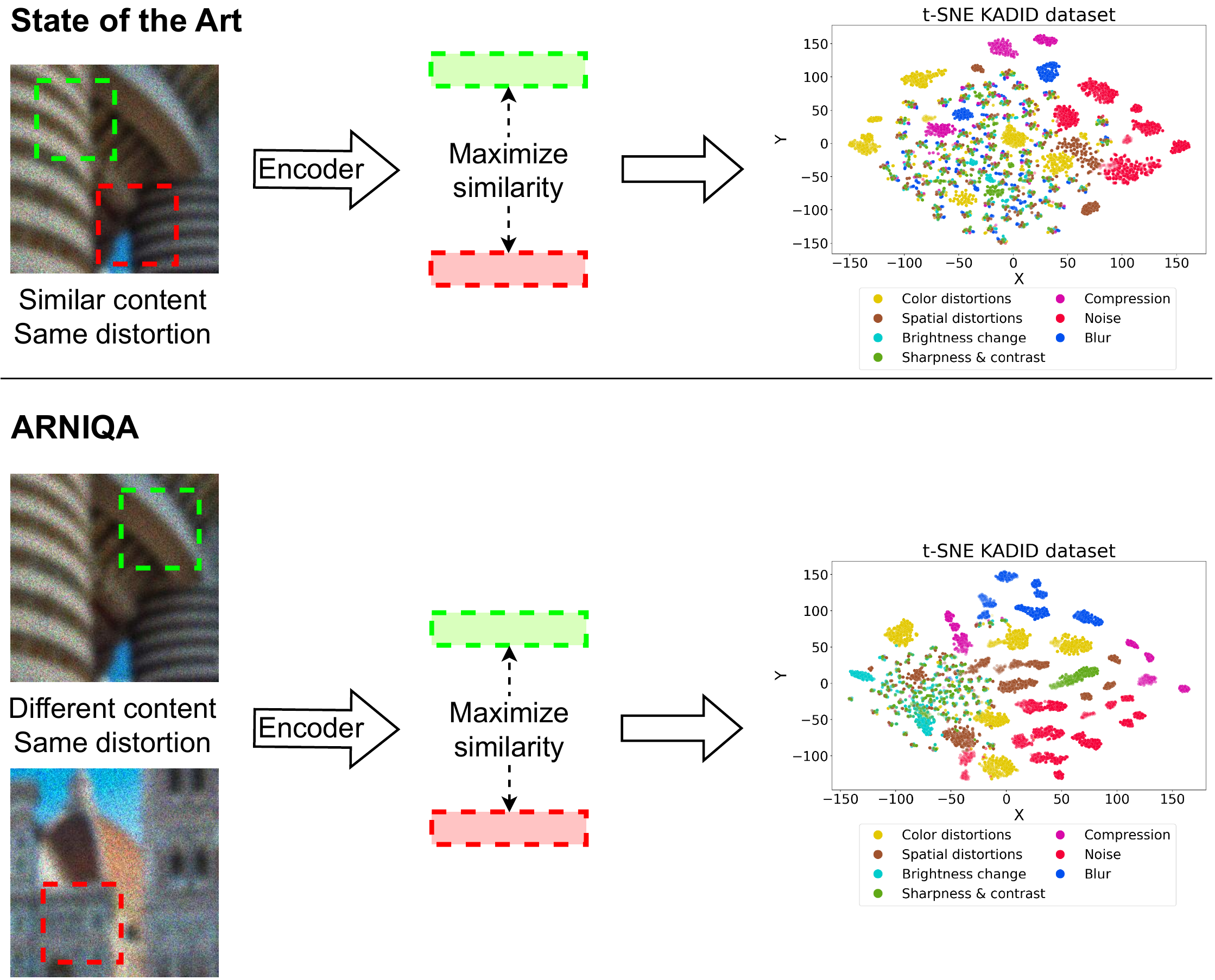}
    \caption{Comparison between our approach and the State of the Art for NR-IQA. While the SotA maximizes the similarity between the representations of crops from the same image, we propose to consider crops from different images degraded equally to learn the image distortion manifold. The t-SNE visualization of the embeddings of the \kadid dataset \cite{kadid10k} shows that, compared to Re-IQA \cite{saha2023re}, \method yields more discernable clusters for different distortions. In the plots, a higher alpha value corresponds to a stronger degradation intensity.}
    \label{fig:teaser}
\end{figure}

Image Quality Assessment (IQA) refers to the computer vision task of automatically evaluating the quality of images with a high correlation with human judgments. Specifically, No-Reference IQA (NR-IQA) focuses on devising methods that can be used when a high-quality reference image is unavailable. NR-IQA finds diverse applications in industries and research domains, including image restoration \cite{wang2021real, liang2021swinir}, captioning \cite{chiu2020assessing}, and multimedia streaming \cite{agnolucci2023perceptual}.

Although supervised learning techniques have shown notable advances in NR-IQA \cite{su2023distortion, su2020blindly, golestaneh2022no, zhang2018blind}, their effectiveness is based on labeled data. Acquiring such annotations is challenging and resource-intensive, given the requirement for a substantial number of ratings to obtain dependable mean opinion scores. For example, the KADID dataset \cite{kadid10k}, which comprises 10125 images synthetically degraded with several distortion types, required approximately 300K annotations. This inherent dependence on labeled data hampers the scalability and broad applicability of supervised approaches.

More recently, several works based on self-supervised learning \cite{chen2020simple, chen2020improved, he2020momentum} have been presented \cite{zhao2023quality, madhusudana2022image, saha2023re}. These methods involve the pre-training of an encoder on unlabeled data with a contrastive loss. Then, the image representations are mapped to the quality scores with a fine-tuning of the encoder weights \cite{zhao2023quality} or by just using a linear regression \cite{madhusudana2022image, saha2023re}. For example, Re-IQA \cite{saha2023re} generates image representations by concatenating low-level and high-level features obtained through a quality-aware and content-aware encoder, respectively. Existing methods involve maximizing the similarity between the representations of two crops of the same distorted image. Therefore, since crops share similar visual information, the model is exposed to content-dependent degradation patterns, which inevitably leads to content-dependent embeddings. The upper part of \cref{fig:teaser} shows the t-SNE visualization \cite{van2008visualizing} of the embeddings of the \kadid \cite{kadid10k} dataset generated by Re-IQA. We notice that the representations related to some types of distortions, \eg blur, are scattered across the space, without being confined into separable clusters. This result stems from the training strategy, as images distorted equally may correspond to different representations due to their diverse content. 

In contrast, we propose a self-supervised approach, named \methodextended, to model the image distortion manifold so that images that exhibit similar degradation patterns correspond to resembling embeddings, despite varying content. We refer to \textit{image distortion manifold} as the continuous space of all the possible degradations to which an image can be subjected. Different regions along this manifold represent various types and degrees of degradation. For example, images showing distinct blur and noise patterns lie in different areas of the manifold. Similarly, images subjected to varying compression rates using the same algorithm correspond to diverse regions within the space. Such a distortion manifold represents image quality in an intrinsic manner. In fact, images that show similar degrees and patterns of degradation are prone to be perceived as having similar quality. At the same time, images exhibiting comparable levels and types of distortion correspond to similar positions in the manifold. Therefore, to map the representation in the manifold to a quality score, it is sufficient to train a simple linear regressor, without the need of fine-tuning the encoder weights. Moreover, by focusing on the inherent distortions within images rather than being dependent on their content, our approach significantly reduces the complexity of the learning process \cite{su2023distortion}. Given two different images that are degraded in the same way, our training strategy consists of extracting a crop from each of them and maximizing the similarity between their representations. At the same time, we maximize the distance from the embeddings of other images degraded in a different manner. In this way, our model learns to recognize image degradation despite varying content. To improve contrastive learning performance, we present a strategy to ensure the presence of hard negative examples within each batch by also considering half-scale images. To train our model, we propose to synthetically distort pristine images with a wide variety of degradations. To this end, we introduce an image degradation model that produces random compositions of consecutively applied distortions. Our degradation model is capable of generating about 100 times more possible distortion compositions than existing approaches. In the lower section of \cref{fig:teaser} we report the t-SNE visualization of the embeddings of the \kadid dataset obtained by \method. We notice that compared to Re-IQA \cite{saha2023re}, our approach produces more easily distinguishable clusters for different types of degradation, thanks to our training strategy.

Extensive experiments show that \method achieves state-of-the-art performance on datasets with both synthetic and authentic (\ie real-world) distortions. Furthermore, since our learning process is less complex, the proposed approach proves to be more data efficient than competing methods, requiring only up to 0.5\% of the training images compared to the competitors. In addition, cross-dataset evaluation and the gMAD competition \cite{ma2016group} demonstrate that \method has better generalization capabilities and is more robust than the baselines.

We summarize our contributions as follows:
\begin{enumerate}
    \item We propose \method, a self-supervised approach for learning the image distortion manifold. By maximizing the similarity between the embeddings of different images distorted equally, we make the encoder generate similar representations for images exhibiting the same degradation patterns regardless of their content;
    \vspace{1ex}
    \item We introduce an image degradation model that randomly assembles ordered sequences of distortions, with $1.9 \cdot 10^{9}$ distinct possible compositions, for synthetically degrading images;
    \vspace{1ex}
    \item \method achieves state-of-the-art performance on NR-IQA datasets with both synthetic and authentic distortions while showing enhanced data efficiency, generalization capabilities, and robustness.
\end{enumerate}

\section{Related Work}

\paragraph{No-Reference Image Quality Assessment}
Due to its importance in both industry and computer vision tasks, No-Reference Image Quality Assessment (NR-IQA) has been an active area of research for several years \cite{mittal2012no, xu2016blind, su2020blindly, ke2021musiq, madhusudana2022image, prabhakaran2023image, saha2023re, wang2023exploring, galteri2022lanbique}. 

Traditional methods \cite{zhang2015feature, mittal2012making, mittal2012no, moorthy2011blind, xue2014blind}, such as BRISQUE \cite{mittal2012no} and NIQE \cite{mittal2012making}, rely on the extraction of handcrafted features from the images. Subsequently, they employ a regression model to predict quality scores. Codebook-based approaches, such as CORNIA \cite{ye2012unsupervised} and HOSA \cite{xu2016blind}, build a visual codebook from local patches to obtain quality-aware features. In recent years, methods using supervised learning achieved a significant boost in performance in NR-IQA \cite{ke2021musiq, su2020blindly, su2023distortion, golestaneh2022no, zhang2018blind, zeng2017probabilistic, ying2020patches}. For example, HyperIQA \cite{su2020blindly} presents a self-adaptive hypernetwork that distinguishes content understanding from quality predictions. The most similar to our work is Su \etal \cite{su2023distortion}, which learns the image distortion manifold in a supervised manner on IQA datasets. Given that it requires distortion-specific information for training, it cannot be used for NR-IQA on datasets with authentic degradations. On the contrary, we model the distortion manifold using unlabeled data with self-supervised learning. Due to their dependence on ground-truth quality scores for training, supervised methods suffer from the scarcity of labeled data for IQA, which are expensive and time-consuming to collect.

Recently, self-supervised learning has emerged as a promising technique for NR-IQA \cite{saha2023re, zhao2023quality, madhusudana2022image}. Self-supervised methods train an encoder on unlabeled data with a contrastive loss and then use its image representations to obtain the final quality scores, either by fine-tuning the model weights \cite{zhao2023quality} or using a linear regressor \cite{madhusudana2022image, zhao2023quality}. QPT \cite{zhao2023quality} proposes a quality-aware contrastive loss based on the assumption that the quality of patches is similar for the same distorted image but differs as the image or the degradations vary. CONTRIQUE \cite{madhusudana2022image} models the representation learning problem as a classification task, where each class is composed of images degraded equally. Re-IQA \cite{saha2023re} trains a quality-aware and a content-aware encoder to generate low-level and high-level representations, respectively. Existing methods are based on maximizing the similarity between the representations of crops of the same distorted image. On the contrary, we maximize the similarity between the embeddings of patches that belong to different images that were degraded equally, regardless of varying content, to model the image distortion manifold. After training, we freeze the encoder weights and map the image representation to the final quality scores with a simple linear regressor.

\paragraph{Image Degradation Models}
Image degradation models aim to synthetically distort images so that the degradation patterns closely resemble those found in real-world scenarios. They play an important role in both blind image restoration \cite{wang2021real, zhang2022closer, zhang2021designing, yue2022blind, shyam2022giqe} and IQA \cite{zhang2018blind, golestaneh2022no, saha2023re, zhao2023quality}. Degradation models differ mainly in how many distinct types of distortion they consider and how they compose them. Specifically, the number of times and the order in which they apply the degradations. RealESRGAN \cite{wang2021real} proposes a second-order degradation model, \ie that performs the distortion process twice but with different parameters. The images are degraded sequentially with one distortion from each of 4 predefined groups, always following the same order. Re-IQA \cite{saha2023re} considers 25 distortion types but applies only one of them to each image, thus not studying combined degradation patterns. QPT \cite{zhao2023quality} presents a second-order degradation model with skip and shuffle operations. It takes into account 3 distortion groups comprising a total of 9 degradation types. In contrast, we introduce an image degradation model that randomly composes ordered sequences of consecutively applied distortions. Given that we consider 24 distortion types divided into 7 groups, we obtain 100 times more possible compositions than existing methods. We rely on our degradation model to synthetically degrade the training images.
\section{Proposed Approach}
Our approach relies on the SimCLR \cite{chen2020simple} framework to train a model composed of a pre-trained ResNet-50 \cite{chen2020simple} encoder and a 2-layer MLP projector that reduces the dimension of the features. We employ unlabeled pristine images distorted with the proposed degradation model for self-supervised learning. After training, we discard the projector and consider the encoder output features as the image representations. Finally, we freeze the encoder and train a linear regressor on top of it to obtain the quality score of an image from its representation.

\subsection{Image Degradation Model}

\begin{figure*}
    \centering
    \includegraphics[width=0.9\linewidth]{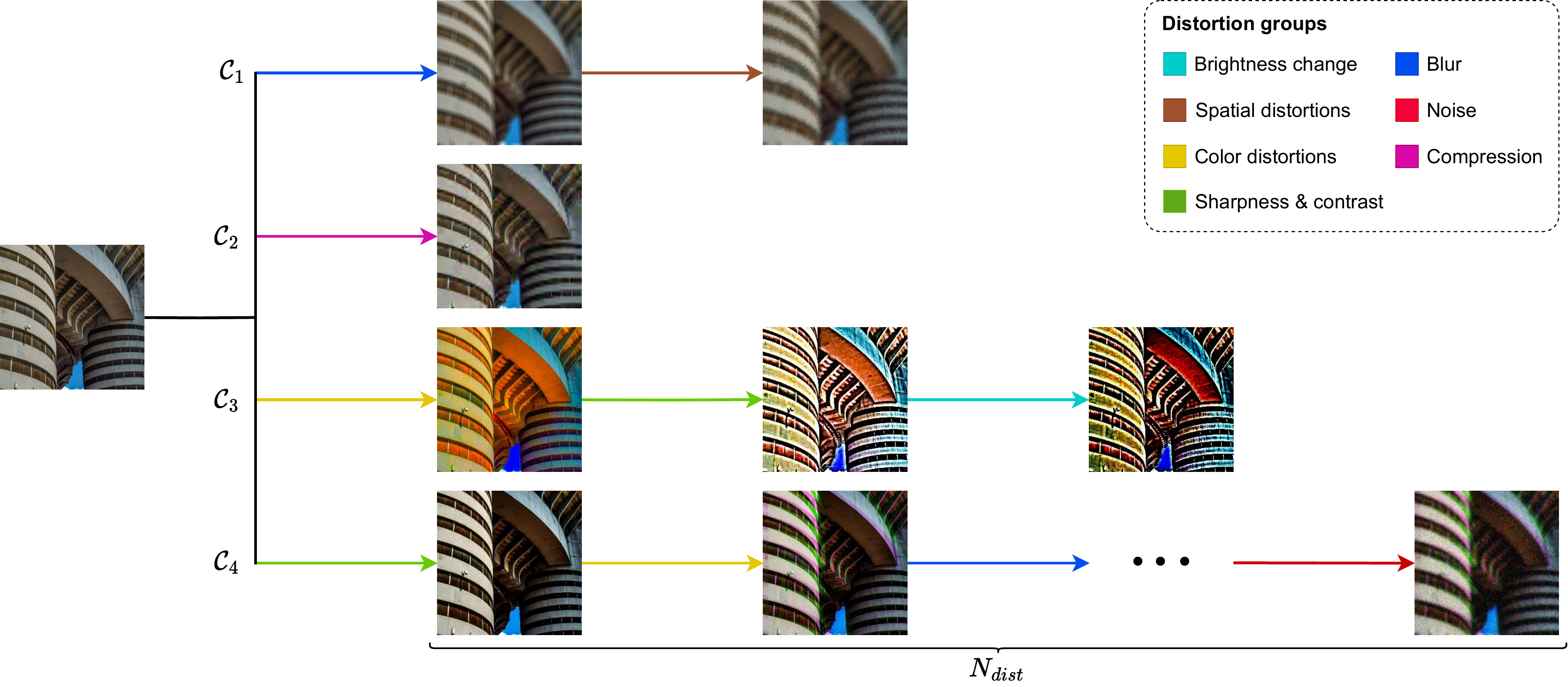}
    \caption{Overview of the proposed image degradation model. We randomly assemble distortion compositions $\mathcal{C}$, \ie ordered sequences of distortions applied consecutively, to synthetically generate images with a wide variety of degradation patterns. Each distortion composition contains a maximum of $\Ndist$ degradations sampled from 7 distinct distortion groups.}
    \label{fig:degradation_model}
\end{figure*}

To effectively learn the image distortion manifold, during training our model must be exposed to a very wide range of diverse degradation patterns. Additionally, it is imperative to possess information about the nature and intensity of degradations within each image for self-supervised learning with a contrastive loss. To address these requirements, we propose to train our model using synthetically degraded images. To this end, we need to make two considerations. First, a broad spectrum of distortion types, spanning varying degrees of intensity, must be taken into account to create a rich collection of degradation patterns. Second, we also need to consider the case of multiple distortions applied at once to investigate how the degradations appear when combined together. Therefore, we introduce an image degradation model that randomly composes ordered sequences of consecutively applied distortions to generate images that exhibit a large variety of degradation patterns. \Cref{fig:degradation_model} shows an overview of the proposed degradation model.

We consider 24 distinct degradation types $D$ divided into the 7 distortion groups $\mathcal{G}\!=\!\{\mathcal{G}_1, \ldots, \mathcal{G}_7\}$ defined by the \kadid \cite{kadid10k} dataset. Each distortion has $L \!=\! 5$ levels of intensity. See \cref{sec:distortion_types} for more details on the specific degradation types. The distortion groups we consider are: 1) Brightness change; 2) Blur; 3) Spatial distortions; 4) Noise; 5) Color distortions; 6) Compression; 7) Sharpness \& contrast. Each of them is defined as $\mathcal{G}_{i} \!=\! \{\ldots, D^{ij}, \ldots \}$, where $i\!\in\!\{1, \ldots, 7 \}$ is the index of the distortion group within $\mathcal{G}$ and $j\!\in\!\{ 1, \ldots, \left | G_{i} \right | \}$ indicates the index of the degradation type within $\mathcal{G}_{i}$, with $\left | \cdot \right |$ that represents the cardinality.

Let $\underline{I}$ be a pristine image. Our aim is to obtain a randomly selected distortion composition $\mathcal{C}$, \ie an ordered sequence of distortions that generates the degraded image $I$ from $\underline{I}$. We define $\Ndist$ as the maximum number of different distortions that can be applied to $\underline{I}$. First, we randomly select a number $\ndist \!=\! \{1, \ldots, \Ndist \}$ of distortions. Then, we sample $\ndist$ distortion types with a uniform distribution from $\ndist$ different degradation groups. In other words, as in \cite{zhang2021designing, zhang2022closer}, for each distortion composition, there can be a maximum of one degradation for each group. Finally, we shuffle the order of the selected distortions and sample a level of intensity for each of them with a given probability distribution. In the end, we obtain a distortion composition $\mathcal{C}\!=\!\{\ldots, D^{ij_{l}}_k, \ldots \}$ where $k\!\in\!\{1, \ldots, \ndist \}$ is the distortion index within $\mathcal{C}$, $i$ and $j$ are defined above and $l\!\in\!\{1, \ldots, L \}$ is the intensity level. Since we want our model to also have access to pristine images during training, we define a hyperparameter $p_{prist}$ and apply the degradation composition to an image with probability $1 - p_{prist}$. Compared to the 9 types of distortion considered by QPT \cite{zhao2023quality}, we take significantly more degradation patterns into account. Moreover, contrary to Re-IQA \cite{saha2023re}, we consecutively apply multiple distortions to the same image, thus studying the effect of their combination.

Applying multiple distortions with a high level of intensity to the same image usually results in a complete disruption of its content. Although our aim encompasses learning areas of the distortion manifold corresponding to very severe degradations, our primary focus resides in regions that are more likely to be related to real-world scenarios. These regions correspond to degradation compositions that alter the content of the image, but not so severely as to make it unrecognizable. In fact, when evaluating the performance of an image restoration model or assessing the quality of a picture uploaded to social platforms, it is unlikely that the images under consideration would be degraded to the point of rendering their content indistinguishable.  Therefore, we propose to sample the intensity level of each distortion with a Gaussian distribution with mean 0 and standard deviation $\sigma$. In this way, lower intensity levels are more likely to be sampled, leading to less severe degradation compositions. Thus, we model regions of the distortion manifold corresponding to degradations most probably corresponding to real-world scenarios in a more fine-grained manner.

Ultimately, our image degradation model is capable of yielding a large variety of distinct distortion compositions. Specifically, the number of possible ways in which the degradations can be assembled is given by:
\begin{equation}
    \sum_{m=1}^{\Ndist}m!L^{m}\left[ \sum_{i=1}^{7}\left | G_{i} \right | \sum_{j=2}^{7}\left | G_{j} \right | \ldots \sum_{k=m}^{7}\left | G_{k} \right | \right]
\end{equation}
As an example, with $\Ndist\!=\!4$, we obtain $1.9 \cdot 10^{9}$ possible compositions, which are about 100 times more than the $2 \cdot 10^{7}$ available with the model proposed in QPT \cite{zhao2023quality}.

\subsection{Training Strategy}

\begin{figure*}
    \centering
    \includegraphics[width=0.9\textwidth]{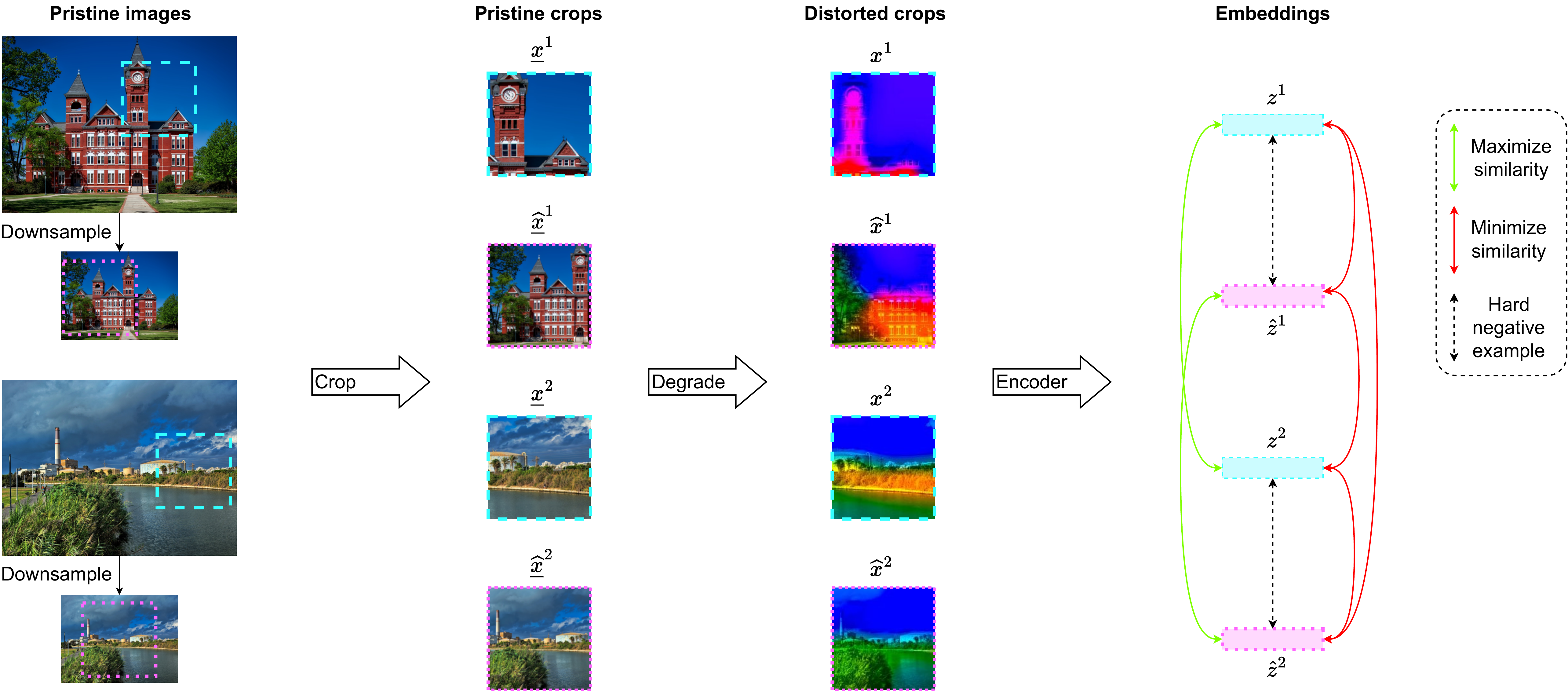}
    \caption{Overview of the proposed training strategy. Given two pristine images, we extract two crops and degrade them equally. Then, we maximize the similarity of their embeddings. At the same time, we minimize the similarity with respect to the embeddings of degraded crops from the half-scale versions of the original images. These embeddings constitute hard negative examples for the representations of the full-scale images since they share similar content and differ only for a downsample distortion. Notice how the original and half-scale degraded crops differ despite being degraded in the same way due to the downsampling operation.}
    \label{fig:method}
\end{figure*}

Existing self-supervised NR-IQA methods, such as Re-IQA \cite{saha2023re}, extract two crops from a single distorted image. Then, they maximize the similarity between their representations. Since the crops share similar visual information, the models learn content-dependent distortion features. In contrast, we maximize the similarity of the representations of crops from two different images with completely diverse content but distorted in the same manner. This way, the encoder learns to model the distortion manifold, thus yielding resembling embeddings for images that exhibit similar degradation patterns, despite varying content. \Cref{fig:method} shows an overview of our training strategy.

Our approach is based on SimCLR \cite{chen2020simple}. SimCLR is a framework for self-supervised learning based on a contrastive loss. Given a training example (\ie an image), SimCLR constructs a positive pair for the contrastive loss by generating two views of the original image with random augmentation techniques. The training process aims to maximize the similarity between the representations of the two views of each training example while maximizing the distance between the embeddings of all the other augmented images in the batch. Thus, the number of examples in each batch is doubled. Intuitively, given that our goal is to learn the image distortion manifold, we can interpret a specific distortion composition as a training example. Therefore, by using it to degrade two different images, we are generating the two views considered by SimCLR. Formally, let $\mathcal{C}\!=\!\{\mathcal{C}_1, \ldots, \mathcal{C}_B\}$ be a batch of distortion compositions obtained with the proposed degradation model, where $B$ is the batch size. Similarly, let $\mathcal{B}^{\scriptscriptstyle 1}\!=\!\{\underline{x}^{\scriptscriptstyle 1}_{\scriptscriptstyle 1}, \ldots, \underline{x}^{\scriptscriptstyle 1}_{\scriptscriptstyle B}\}$ and $\mathcal{B}^{\scriptscriptstyle 2}\!=\!\{\underline{x}^{\scriptscriptstyle 2}_{\scriptscriptstyle 1}, \ldots, \underline{x}^{\scriptscriptstyle 2}_{\scriptscriptstyle B}\}$ be two batches of pristine images randomly selected from the training dataset. For each pair $(\underline{x}^{\scriptscriptstyle 1}_i, \underline{x}^{\scriptscriptstyle 2}_i)$ where $i\! \in \! \{1, \ldots, B\}$, we extract a random crop from each image and employ $\mathcal{C}_i$ to obtain the degraded version $(x^{\scriptscriptstyle 1}_{i}, x^{\scriptscriptstyle 2}_{i})$. Each pair constitutes the two views of the SimCLR framework, and their embeddings represent a positive pair in the contrastive loss. 

However, since the proposed degradation model has a very large number of possible compositions, the given batch of distortion compositions $\mathcal{C}$ could lead to considerably different image pairs. In that case, it would be trivial for the model to discriminate between the different examples, making the learning process less effective. To avoid this issue, we propose a strategy to ensure the presence of hard negative examples in each batch, which is known to enhance contrastive learning \cite{robinson2020contrastive, kalantidis2020hard}. Given an image pair $(\underline{x}^{\scriptscriptstyle 1}_i, \underline{x}^{\scriptscriptstyle 2}_i)$ defined as above, we downsample the images to half size before cropping, resulting in $(\wh{\underline{x}}^{\scriptscriptstyle 1}_i, \wh{\underline{x}}^{\scriptscriptstyle 2}_i)$. After applying $\mathcal{C}_i$, we obtain the distorted image pair $(\wh{x}^{\scriptscriptstyle 1}_i, \wh{x}^{\scriptscriptstyle 2}_i)$. Given that the downsampling operation fundamentally reduces the pixel count, it inherently results in information loss and thus can be viewed as a degradation. Therefore, this process can be likened to prepending a downsampling degradation to each distortion composition $\mathcal{C}_i$. Finally, we apply this technique to all the image pairs and add the new $B$ pairs to the batch, thereby doubling the batch size and the number of negative examples, which improves the performance of contrastive learning \cite{chen2020simple, grill2020bootstrap}. Moreover, since we use all the images both at full-scale and half-scale, the size of the training dataset is also doubled. Thanks to our strategy, the model always has to discriminate between the two examples $x^{\scriptscriptstyle 1}_{i}$ and $\wh{x}^{\scriptscriptstyle 1}_i$, which mutually serve as hard negatives for each other. Indeed, they share similar content as they are crops taken from the same image at two different scales, and their degradation differs only for a downsample distortion. Therefore, by minimizing the similarity between the representations of $x^{\scriptscriptstyle 1}_{i}$ and $\wh{x}^{\scriptscriptstyle 1}_i$, our model learns to discriminate between images with slightly different degradation, even if they share similar content. The same considerations apply for $x^{\scriptscriptstyle 2}_{i}$ and $\wh{x}^{\scriptscriptstyle 2}_i$. CONTRIQUE \cite{madhusudana2022image} considers images at half-scale as well but regards them as positive examples belonging to the same distortion class. On the contrary, we treat the half-scale resolution crops as challenging negative examples, since they actually differ for a single distortion.

Formally, let $f(\cdot)$ be the ResNet-50 encoder and $g(\cdot)$ the 2-layer MLP projector that we use for dimensionality reduction. Then, given an image $x \! \in \! \mathbb{R}^{3 \times H \times W}$, with $H$ and $W$ representing respectively its height and width, we compute its representation $z$ with:
\begin{align}
    h &= f(x) \in \mathbb{R}^C, & z = g(h) &= g(f(x)) \in \mathbb{R}^D
\end{align}
where $C$ and $D$ are the number of channels of the encoder and the projector, respectively. First, we generate the embeddings of all the views, both at full- and half-scale. Then, following SimCLR, we employ the NT-Xent contrastive loss \cite{chen2020simple} for training. To this end, we define the loss terms:
\begin{equation}
    \resizebox{0.8\columnwidth}{!}{$ \begin{aligned}
    \ell^{\scriptscriptstyle 1, 2}_{i} &= -\log \frac{\gamma^{\scriptscriptstyle 1, 2}_{i, i}}{\sum\limits_{k = 1}^{B} \left [ \gamma^{\scriptscriptstyle 1, 2}_{i, k} + \gamma^{\scriptscriptstyle 1, 2}_{i, \wh{k}} + \gamma^{\scriptscriptstyle 1, 1}_{i, \wh{k}} \right ] + \sum\limits_{k \ne i}^{B}  \gamma^{\scriptscriptstyle 1, 1}_{i, k}} \\
    \wh{\ell}^{\scriptscriptstyle 2, 1}_{i} &= -\log \frac{\gamma^{\scriptscriptstyle 2, 1}_{\wh{i}, \wh{i}}}{\sum\limits_{k = 1}^{B} \left [ \gamma^{\scriptscriptstyle 2, 1}_{\wh{i}, \wh{k}} + \gamma^{\scriptscriptstyle 2, 1}_{\wh{i}, k} + \gamma^{\scriptscriptstyle 2, 2}_{\wh{i}, k} \right ] + \sum\limits_{k \ne i}^{B}  \gamma^{\scriptscriptstyle 2, 2}_{\wh{i}, \wh{k}}}
    \end{aligned}$}
    \label{eq:term_loss_info_nce}
\end{equation}
where $\gamma^{\scriptscriptstyle 1, 2}_{i, \wh{k}} = e^{( cos(z_i^1, \wh{z}_k^2) / \tau )}$ and $cos(\cdot)$ and $\tau$ represent the cosine similarity and a temperature hyperparameter, respectively. Hence, the overall training loss is given by:
\begin{equation}
    \mathcal{L} = \frac{1}{4B} \sum_{i=1}^{B} \left [ \ell^{1, 2}_{i} + \ell^{2, 1}_{i} + \wh{\ell}^{1, 2}_{i} + \wh{\ell}^{2, 1}_{i} \right ]
    \label{eq:total_loss_info_nce}
\end{equation}
Intuitively, the loss maximizes the similarity between the representation of each view and the corresponding one, while maximizing the distance to all the other views, both at full-scale and half-scale. Therefore, we consider 2 (views) $\times$ 2 (scales) $\times$ $B$ (batch size) $\!=\!$ 4$B$ elements in total.

After training, our model has learned a distortion manifold and hence generates similar embeddings for images degraded in the same way, regardless of their content.
\section{Experimental Results}

\subsection{Implementation Details}
We train our model for 10 epochs using a stochastic gradient descent (SGD) optimizer with momentum 0.9 and weight decay $1e-4$. Starting from a learning rate of $1e-3$, we employ a cosine annealing with warm restarts scheduler \cite{loshchilov2016sgdr}. Differently from \cite{madhusudana2022image, saha2023re}, we start from a pre-trained ResNet-50 encoder and fine-tune its weights during training. The encoder and the projector have a number of channels $C$ and $D$ of 2048 and 128, respectively. During training, we use a patch size of 224, a temperature $\tau$ of 0.1, and a batch size of 16. Regarding the image degradation model, we set the maximum number of distortions $\Ndist$ to 4, the probability of using pristine images $p_{prist}$ to 0.05, and the standard deviation of the Gaussian distribution $\sigma$ to 2.5. 

\subsection{Datasets}
We employ the 140K pristine images from the KADIS dataset \cite{kadid10k} for training, discarding the 700K degraded ones it provides. In fact, we use our image degradation model to obtain the degraded versions of the pristine images.

We test \method on datasets with both synthetic and authentic distortions. These consist of collections of degraded images labeled with subjective opinions of picture quality in the form of Mean Opinion Score (MOS). We consider four synthetically degraded datasets: \live \cite{sheikh2006statistical}, \csiq \cite{larson2010most}, \tid \cite{ponomarenko2013color}, and \kadid \cite{kadid10k}.  \live comprises 779 images degraded with 5 types of distortion at 5 levels of intensity, with 29 reference images as the base. \csiq, on the other hand, stems from 30 reference images, each undergoing 6 types of distortions at 5 levels of intensity, yielding 866 images. \tid and \kadid contain 3000 and 10125 images synthetically degraded with 24 and 25 types of distortion at 5 different levels of intensity, starting from 25 and 81 reference images, respectively. Regarding the datasets with authentic distortions, we consider \flive \cite{ying2020patches} and \spaq \cite{fang2020perceptual}. \flive is the largest existing dataset for NR-IQA, comprising nearly 40K real-world images. \spaq contains 11K high-resolution images captured by several mobile devices. Similar to \cite{fang2020perceptual, madhusudana2022image}, for evaluation, we resize the \spaq images so that the shorter side is 512. 

\begin{table*}
  \centering
  \Large
  \resizebox{\linewidth}{!}{ 
  \begin{tabular}{lccccccccccccccc} 
  \toprule
  \multicolumn{1}{c}{} & \multicolumn{1}{c}{} & \multicolumn{8}{c}{Synthetic Distortions} & \multicolumn{4}{c}{Authentic Distortions} & \multicolumn{2}{c}{} \\
  \cmidrule(lr){3-10}
  \cmidrule(lr){11-14}
  \multicolumn{1}{c}{} & \multicolumn{1}{c}{} & \multicolumn{2}{c}{\live} & \multicolumn{2}{c}{\csiq} & \multicolumn{2}{c}{\tid} &\multicolumn{2}{c}{\kadid} & \multicolumn{2}{c}{\flive} & \multicolumn{2}{c}{\spaq} & \multicolumn{2}{c}{Average} \\
  \multicolumn{1}{l}{Method} & \multicolumn{1}{c}{Type} & \srcc & \plcc & \srcc & \plcc & \srcc & \plcc & \srcc & \plcc & \srcc & \plcc & \srcc & \plcc & \srcc & \plcc \\ 
  \midrule
  BRISQUE \cite{mittal2012no} & \multirow{2}{*}{Handcrafted} & 0.939 & 0.935 & 0.746 & 0.829 & 0.604 & 0.694 & 0.528 & 0.567 & 0.288 & 0.373 & 0.809 & 0.817 & 0.652 & 0.703 \\
  NIQE \cite{mittal2012making} &  & 0.907 & 0.901 & 0.627 & 0.712 & 0.315 & 0.393 & 0.374 & 0.428 & 0.211 & 0.288 & 0.700 & 0.709 & 0.522 & 0.572 \\ \midrule[0.01em]
  CORNIA \cite{ye2012unsupervised} & \multirow{2}{*}{Codebook} & 0.947 & 0.950 & 0.678 & 0.776 & 0.678 & 0.768 & 0.516 & 0.558 & -- & -- & 0.709 & 0.725 & -- & -- \\
  HOSA \cite{xu2016blind} &  & 0.946 & 0.950 & 0.741 & 0.823 & 0.735 & 0.815 & 0.618 & 0.653 & -- & -- & 0.846 & 0.852 & -- & -- \\ \midrule[0.01em]
  DB-CNN \cite{zhang2018blind} & \multirow{4}{*}{\shortstack[c]{Supervised \\ learning}} & 0.968 & \underline{0.971} & 0.946 & 0.959 & 0.816 & 0.865 & 0.851 & 0.856 & 0.554 & 0.652 & 0.911 & 0.915 & 0.841 & 0.870 \\
  HyperIQA \cite{su2020blindly} &  & 0.962 & 0.966 & 0.923 & 0.942 & 0.840 & 0.858 & 0.852 & 0.845 & 0.535 & 0.623 & \underline{0.916} & \underline{0.919} & 0.838 & 0.859 \\
  TReS \cite{golestaneh2022no} &  & 0.969 & 0.968 & 0.922 & 0.942 & \underline{0.863} & \underline{0.883} & 0.859 & 0.858 & 0.554 & 0.625 & -- & -- & -- & -- \\
  Su \etal \cite{su2023distortion} &  & \textbf{0.973} & \textbf{0.974} & 0.935 & 0.952 & 0.815 & 0.859 & 0.866 & 0.874 & -- & -- & -- & -- & -- & -- \\ \midrule[0.01em]
  CONTRIQUE \cite{madhusudana2022image} & \multirow{2}{*}{SSL + LR} & 0.960 & 0.961 & 0.942 & 0.955 & 0.843 & 0.857 & \textbf{0.934} & \textbf{0.937} & 0.580 & 0.641 & 0.914 & \underline{0.919} & \underline{0.862} & 0.878 \\
  Re-IQA \cite{saha2023re} &  & \underline{0.970} & \underline{0.971} & \underline{0.947} & \underline{0.960} & 0.804 & 0.861 & 0.872 & 0.885 & \textbf{0.645} & \textbf{0.733} & \textbf{0.918} & \textbf{0.925} & 0.859 & \underline{0.889} \\ \midrule
  \rowcolor{tabhighlight}
  \textbf{\method} & SSL + LR & 0.966 & 0.970 & \textbf{0.962} & \textbf{0.973} & \textbf{0.880} & \textbf{0.901} & \underline{0.908} & \underline{0.912} & \underline{0.595} & \underline{0.671} & 0.905 & 0.910 & \textbf{0.869} & \textbf{0.890} \\
  \bottomrule
  \end{tabular}}
  \caption{Comparison between the proposed approach and competing methods on datasets with synthetic and authentic distortions. Best and second-best scores are highlighted in bold and underlined, respectively. -- denotes results not reported in the original paper. SSL and LR stands for self-supervised learning and linear regression, respectively.}
  \label{tab:main_results}
\end{table*}

\subsection{Evaluation Protocol}
To evaluate the performance, we employ Spearman's rank order correlation coefficient (\srcc) and Pearson's linear correlation coefficient (\plcc) to measure prediction monotonicity and accuracy, respectively.

Following \cite{madhusudana2022image,saha2023re}, we randomly divide the datasets into 70\%, 10\%, and 20\% splits corresponding to training, validation, and test sets, respectively. Splits are selected based on reference images to ensure no overlap of contents. We employ the ground-truth MOS scores of the training split to train a Ridge regressor \cite{hoerl1970ridge} with an L2 loss. Note that we do not perform any fine-tuning of the encoder weights during the evaluation. Similarly to \cite{madhusudana2022image, saha2023re}, we use the validation split to identify the regularization coefficient of the regressor via a grid search over values in the range $\left[ 10^{-3}, 10^{3} \right]$.  During testing, we compute the image features at full-scale and half-scale and concatenate them to obtain the final representation, as in \cite{madhusudana2022image}. Similarly to \cite{zhao2023quality}, we take the four corners and the center crops at both scales and average the corresponding predicted quality scores to obtain the final one. To remove any bias in the selection of the training set, we repeat the training/test procedure 10 times and report the median results. Given the large size of the dataset, for FLIVE we use only the official train-test split \cite{ying2020patches}.

\subsection{Results}

\begin{table}
    \centering
    \Large
    \resizebox{\linewidth}{!}{ 
    \begin{tabular}{llcccch}
        \toprule
        \multicolumn{1}{c}{} & \multicolumn{1}{c}{} & \multicolumn{5}{c}{Method} \\
        \cmidrule(lr){3-7}
        Training & Testing & HyperIQA & Su \etal & CONTRIQUE$^{\dagger}$ & Re-IQA$^{\dagger}$ & \textbf{\method} \\
        \midrule
        \live & \csiq & 0.744 & 0.777 & 0.803 & 0.795 & \textbf{0.904} \\
        \live & \tid & 0.541 & 0.561 & 0.640 & 0.588 & \textbf{0.697} \\
        \live & \kadid & 0.492 & 0.506 & 0.699 & 0.557 & \textbf{0.764} \\
        \csiq & \live & 0.926 & \textbf{0.930} & 0.912 & 0.919 & 0.921 \\
        \csiq & \tid & 0.541 & 0.550 & 0.570 & 0.575 & \textbf{0.721} \\
        \csiq & \kadid & 0.509 & 0.515 & 0.696 & 0.521 & \textbf{0.735} \\
        \tid & \live & 0.876 & 0.892 & \textbf{0.904} & 0.900 & 0.869 \\
        \tid & \csiq & 0.709 & 0.754 & 0.811 & 0.850 & \textbf{0.866} \\
        \tid & \kadid & 0.581 & 0.554 & 0.640 & 0.636 & \textbf{0.726} \\
        \kadid & \live & \textbf{0.908} & 0.896 & 0.900 & 0.892 & 0.898 \\
        \kadid & \csiq & 0.809 & 0.828 & 0.773 & 0.855 & \textbf{0.882} \\
        \kadid & \tid & 0.706 & 0.687 & 0.612 & \textbf{0.777} & 0.760 \\
        \bottomrule
    \end{tabular}}
    \caption{Cross-dataset evaluation results for the \srcc metric. $^{\dagger}$ denotes results evaluated by us with the official pre-trained models. Best scores are highlighted in bold.}
    \label{tab:cross_dataset}
\end{table}

In \cref{tab:main_results} we compare the performance of the proposed approach with other state-of-the-art methods. \method achieves competitive performance for both synthetic and authentic distortions and obtains the best results on average. In particular, we notice how our method outperforms Su \etal \cite{su2023distortion}, which also aims to learn the distortion manifold but in a supervised manner and directly on IQA datasets. Furthermore, contrary to our approach, Su \etal cannot be evaluated on datasets with authentic degradations, as it requires distortion-specific information for training. Compared to other self-supervised approaches, namely CONTRIQUE \cite{madhusudana2022image} and Re-IQA \cite{saha2023re}, \method achieves comparable or better performance. However, our method is significantly more data-efficient than the competitors. Indeed, we employ 140K (training dataset) $\times$ 2 (scales) $\times$ 10 (epochs) $\!=\!$ 2.8M images for training. In contrast, doing similar computations, we get that CONTRIQUE uses 65M images, while Re-IQA requires 512M and 38M images for the content and quality encoders, respectively. See \cref{sec:analysis_data_efficiency} for more details. Therefore, \method achieves state-of-the-art performance while requiring only up to 0.5\% of the data compared to competing methods. The reason is that focusing solely on the degradation patterns within images reduces the complexity of the learning process compared to depending on image content as well, as also observed by \cite{su2023distortion}.

We evaluate the generalization capabilities of our model by measuring cross-dataset performance. We train the regressor on the whole training dataset and then use it to obtain the quality predictions on the testing dataset. We report the results for the \srcc metric in \cref{tab:cross_dataset}. \method significantly outperforms all the competing methods. In particular, the proposed approach achieves the largest improvements compared to the baselines when training on a dataset comprising few distortion types (\eg \csiq) and testing on one with a large variety of different degradations (\eg \tid). We hypothesize that the reason behind this outcome is that our method makes the encoder model the regions of the distortion manifold that correspond to distinct types of degradations in a consistent way. In other words, the mapping from the distortion manifold to the quality scores is consistent across different types of degradation. Therefore, a regressor trained by mapping only some specific regions of the manifold -- \ie considering only a few different distortions -- to quality scores behaves well even when used on unseen degradation types. We will study this phenomenon more thoroughly in future work. 

\begin{figure}
    \centering
    \begin{subfigure}{0.225\linewidth}
        \includegraphics[width=\linewidth]{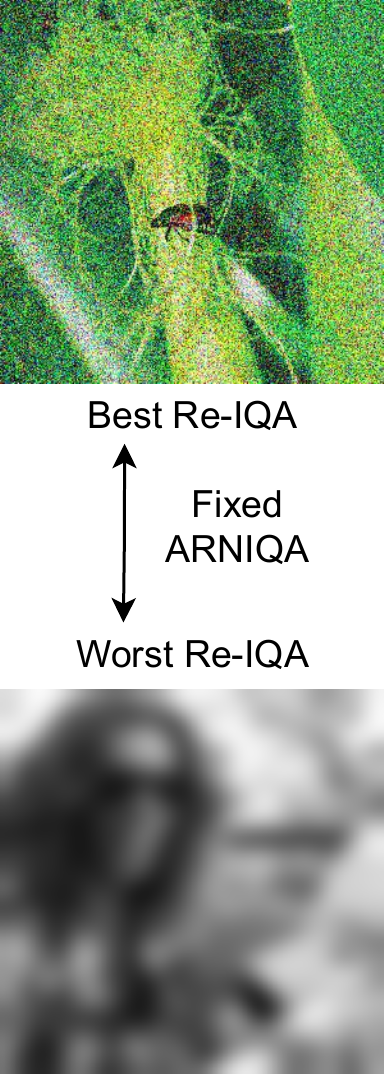}
        \caption{}
        \label{fig:ours_def_low}
    \end{subfigure}
    \hfill
    \begin{subfigure}{0.225\linewidth}
        \includegraphics[width=\linewidth]{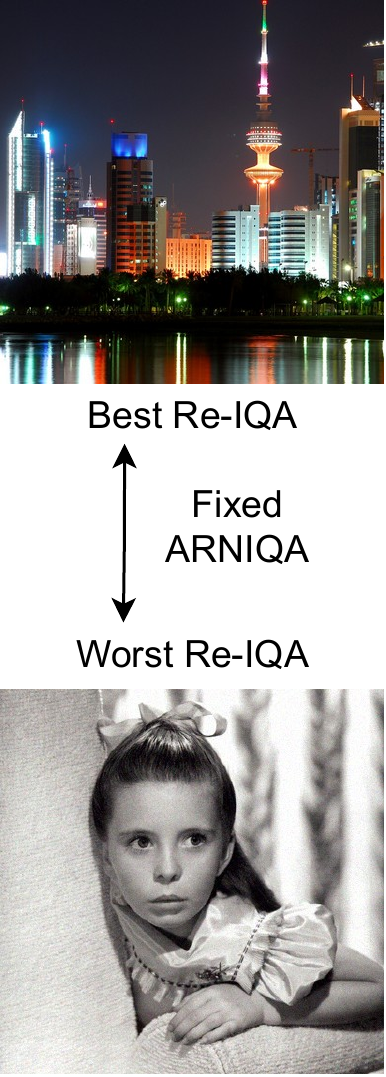}
        \caption{}
        \label{fig:ours_def_high}
    \end{subfigure}
    \hfill
    \begin{subfigure}{0.225\linewidth}
        \includegraphics[width=\linewidth]{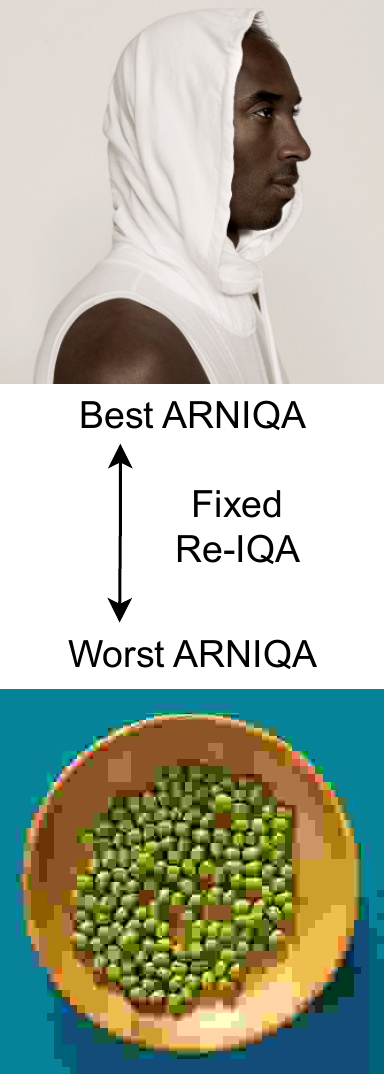}
        \caption{}
        \label{fig:ours_att_low}
    \end{subfigure}
    \hfill
    \begin{subfigure}{0.225\linewidth}
        \includegraphics[width=\linewidth]{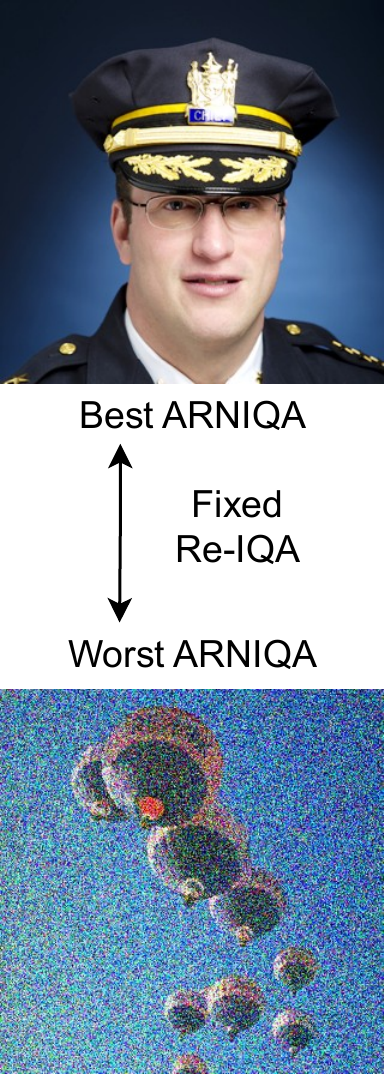}
        \caption{}
        \label{fig:ours_att_high}
    \end{subfigure}
    \caption{gMAD competition results between \method and Re-IQA \cite{saha2023re}. (a) and (b): Fixed \method at a low- and high-quality level, respectively. (c) and (d): Fixed Re-IQA at a low- and high-quality level, respectively.}
    \label{fig:gmad_reiqa}
\end{figure}

To evaluate the robustness of the model, we conduct the group maximum differentiation (gMAD) competition \cite{ma2016group} between \method and Re-IQA on the Waterloo Exploration Database \cite{ma2016waterloo}, a dataset with synthetically degraded images without MOS annotations. We fix one model to act as the defender and we group its quality predictions into several levels. Then, the other model functions as the attacker by identifying the image pairs within each level that differ the most in terms of quality. Therefore, for a model to be robust, the selected image pairs should exhibit similar quality when functioning as the defender and show a noticeable quality difference when acting as the attacker. We show the results in \cref{fig:gmad_reiqa}. When we fix \method (\cref{fig:ours_def_low,fig:ours_def_high}), Re-IQA is unable to identify image pairs showing an obvious quality difference. On the contrary, when \method acts as the attacker (\cref{fig:ours_att_low,fig:ours_att_high}), it manages to spot the failures of Re-IQA by finding image pairs that clearly have significantly different quality. Thus, the proposed approach proves to be more robust than Re-IQA.

We provide additional experimental results in \cref{sec:additional_experimental_results}.

\subsection{Ablation Studies}

\paragraph{Image Degradation Model} We conduct ablation studies on our image degradation model: 1) \textit{RealESRGAN}: we replicate the degradation model of RealESRGAN \cite{wang2021real}; 2) \textit{w/o gaussian}: we sample the intensity level of each degradation with a uniform distribution instead of a Gaussian one;
3) \textit{$\Ndist\!=\!1$} and 4) \textit{$\Ndist\!=\!7$}: we reduce and increase the maximum number of distorsions $\Ndist$, respectively.

The upper part of \cref{tab:ablation} shows the results for the \srcc metric. 
We observe that the RealESRGAN degradation model obtains poor performance on all the datasets. Expectedly, considering only 4 distortion groups and applying them always in the same order limits the variety of degradation patterns the model is exposed to during training, hampering the learning process. We notice that sampling the levels of intensity of the degradations with a uniform distribution degrades the performance compared to using a Gaussian one. This is because it leads to more coarse modeling of the regions of the manifold corresponding to degradations more likely to be related to those found in real-world scenarios. Finally, the variants of the degradation model with \textit{$\Ndist\!=\!1$} and \textit{$\Ndist\!=\!7$} generate images that respectively contain no combined degradation patterns and too strong distortions, thus hampering the training process.

\begin{table}[t]
    \centering
    \Large
    \resizebox{\linewidth}{!}{ 
    \begin{tabular}{lccccc}
        \toprule
        Method & \live & \csiq & \tid & \kadid & Average\\
        \midrule
        RealESRGAN & 0.926 & 0.896 & 0.616 & 0.727 & 0.791 \\
        w/o gaussian & 0.965 & 0.953 & 0.866 & \textbf{0.920} & 0.926 \\
        $\Ndist\!=\!1$ & 0.966 & 0.957 & 0.857 & 0.916 & 0.924 \\
        $\Ndist \!=\!7$ & \textbf{0.970} & 0.957 & 0.868 & 0.902 & 0.924 \\ \midrule[.01em]
        same image & 0.940 & 0.863 & 0.721 & 0.775 & 0.825 \\
        w/o HN & 0.966 & 0.960 & 0.851 & 0.908 & 0.921 \\
        \midrule
        \rowcolor{tabhighlight}
        \textbf{\method} & 0.966 & \textbf{0.962} & \textbf{0.880} & 0.908 & \textbf{0.929} \\
        \bottomrule
    \end{tabular}}
    \caption{Ablation studies results for the \srcc metric. Best scores are highlighted in bold.}
    \label{tab:ablation}
\end{table}

\paragraph{Training Strategy} We perform ablation studies on our training strategy: 1) \textit{same image}: we extract the crops from the same image, instead of from two different ones; 2) \textit{w/o HN}: we do not employ our strategy to obtain hard negative examples and, for a fair comparison, we double the batch size to have the same number of negative examples. 

We report the results for the \srcc metric in the lower section of \cref{tab:ablation}. We observe that extracting two crops from the same degraded image leads to poor performance. Even if it proved to be a viable technique to achieve state-of-the-art results in NR-IQA, it requires more convoluted approaches compared to ours, such as considering multiple loss terms \cite{zhao2023quality} or two different encoders \cite{saha2023re}. Furthermore, we notice that our strategy to guarantee the presence of hard negatives in every batch improves the results compared to using the same number of randomly sampled negative examples, as expected for contrastive learning \cite{robinson2020contrastive, kalantidis2020hard}. 
\section{Conclusion}
In this work, we present a self-supervised approach, named \method, to learn the image distortion manifold for NR-IQA. First, we introduce an image degradation model that randomly assembles ordered sequences of distortions, with about 100 times more possible compositions than competing methods. Second, we propose a training strategy that maximizes the similarity between the embeddings of crops belonging to distinct images degraded equally, regardless of their content. This way, we model the distortion manifold so that a simple linear regressor can effectively map image representations to quality scores. The experiments show that \method achieves state-of-the-art performance on datasets with both synthetic and authentic distortions. Also, our method exhibits enhanced generalization capabilities, data efficiency, and robustness compared to the baselines. In future work, we will study how our learned distortion manifold can be used for blind image restoration.

\paragraph{Acknowledgments}
This work was partially supported by the European Commission under European Horizon 2020 Programme, grant number 101004545 - ReInHerit.
{
    \small
    \bibliographystyle{ieeenat_fullname}
    \bibliography{main}
}

\clearpage
\maketitlesupplementary

\section{Analysis on Data Efficiency} \label{sec:analysis_data_efficiency}
In Sec. {\color{red}4.4} we show that \method achieves state-of-the-art performance on several IQA datasets with both synthetic and authentic distortions. In addition, our approach proves to be more data-efficient than competing self-supervised methods, since it requires fewer training examples.

We recall that we rely on the 140K pristine images from the KADIS dataset \cite{kadid10k} synthetically distorted with our degradation model to train our model. Given that we consider images both at full-scale and half-scale (see Sec. {\color{red}3.2}), we double the size of the training dataset. For all of the experiments, we train our model for 10 epochs. Therefore, for training, we use a total of 140K (training dataset) $\times$ 2 (scales) $\times$ 10 (epochs) $\!=\!$ 2.8M images. 

In contrast, CONTRIQUE \cite{madhusudana2022image} considers a combination of images with synthetic and authentic distortions for training, for a total of 1.3M. Specifically, the authors use the 700K synthetically distorted images from the KADIS dataset and the union of 4 datasets with realistic distortions: 255K images from AVA \cite{murray2012ava}, 330K images from COCO \cite{lin2014microsoft}, 2450 images from CERTH-Blur \cite{mavridaki2014no}, 33K images from VOC \cite{everingham2010pascal}. The CONTRIQUE model employs both full-scale and half-scale images and is trained for 25 epochs. Therefore, the total number of training examples required by CONTRIQUE is given by 1.3M (training dataset) $\times$ 2 (scales) $\times$ 25 (epochs) $\!=\!$ 65M. 

Instead, Re-IQA uses two different datasets, both at full-scale and half-scale, as well as a diverse number of epochs, for the content-aware and the quality-aware encoder. In particular, the authors train the content-aware encoder on the 1.28 images of the ImageNet dataset \cite{deng2009imagenet} for 200 epochs. Thus, the total number of training examples for the content-aware encoder is given by 1.28M (training dataset) $\times$ 2 (scales) $\times$ 200 (epochs) $\!=\!$ 512M. For the quality-aware encoder, Re-IQA uses the 140K pristine images from the KADIS dataset and the same combination of datasets with authentic distortions as CONTRIQUE, for a total of 760K images. Given that the authors train the quality-aware encoder for 25 epochs, the total number of training examples results in 760K (training dataset) $\times$ 2 (scales) $\times$ 25 (epochs) $\!=\!$ 38M. Considering both the content-aware and the quality-aware encoders, Re-IQA requires a total of 550M images for training.

Ultimately, despite using only the 4.3\% and 0.5\% of the training examples compared, respectively, to CONTRIQUE and Re-IQA, \method manages to achieve state-of-the-art performance on several IQA datasets, thereby showing improved data efficiency.

\section{Additional Experimental Results} \label{sec:additional_experimental_results}

\subsection{Full-Reference Image Quality Assessment}

\begin{table*}
  \centering
  \Large
  \resizebox{0.6\linewidth}{!}{ 
  \begin{tabular}{lccccccccc} 
  \toprule
  \multicolumn{1}{c}{} & \multicolumn{1}{c}{} & \multicolumn{2}{c}{\live} & \multicolumn{2}{c}{\csiq} & \multicolumn{2}{c}{\tid} &\multicolumn{2}{c}{\kadid} \\
  \multicolumn{1}{l}{Method} & \multicolumn{1}{c}{Type} & \srcc & \plcc & \srcc & \plcc & \srcc & \plcc & \srcc & \plcc \\ 
  \midrule
  PSNR & \multirow{4}{*}{Traditional} & 0.881 & 0.868 & 0.820 & 0.824 & 0.643 & 0.675 & 0.677 & 0.680 \\
  SSIM \cite{wang2004image} &  & 0.921 & 0.911 & 0.854 & 0.835 & 0.642 & 0.698 & 0.641 & 0.633 \\ 
  FSIM \cite{zhang2011fsim} &  & 0.964 & 0.954 & 0.934 & 0.919 & 0.852 & 0.875 & 0.854 & 0.850 \\
  VSI \cite{zhang2014vsi} &  & 0.951 & 0.940 & 0.944 & 0.929 & 0.902 & 0.903 & 0.880 & 0.878 \\ \midrule[0.01em]
  PieAPP \cite{prashnani2018pieapp} & \multirow{4}{*}{\shortstack[c]{Deep \\ learning}} & 0.915 & 0.905 & 0.900 & 0.881 & 0.877 & 0.850 & 0.869 & 0.869 \\
  LPIPS \cite{zhang2018unreasonable} &  & 0.932 & 0.936 & 0.884 & 0.906 & 0.673 & 0.756 & 0.721 & 0.713 \\
  DISTS \cite{ding2020image} &  & 0.953 & 0.954 & 0.942 & 0.942 & 0.853 & 0.873 & -- & -- \\
  DRF-IQA \cite{kim2020dynamic} &  & \textbf{0.983} & \textbf{0.983} & \underline{0.964} & 0.960 & \textbf{0.944} & \textbf{0.942} & -- & -- \\ \midrule[0.01em]
  CONTRIQUE-FR \cite{madhusudana2022image} & \multirow{2}{*}{SSL + LR} & 0.966 & 0.966 & 0.956 & \underline{0.964} & 0.909 & 0.915 & \textbf{0.946} & \textbf{0.947} \\
  Re-IQA-FR \cite{saha2023re} &  & \underline{0.969} & \underline{0.974} & 0.961 & 0.962 & \underline{0.920} & \underline{0.921} & \underline{0.933} & \underline{0.936} \\ \midrule
  \rowcolor{tabhighlight}
  \textbf{\method-FR} & SSL + LR & \underline{0.969} & 0.972 & \textbf{0.971} & \textbf{0.975} & 0.898 & 0.901 & 0.920 & 0.919 \\
  \bottomrule
  \end{tabular}}
  \caption{Comparison between the proposed approach and competing methods for the FR-IQA task. Best and second-best scores are highlighted in bold and underlined, respectively, -- denotes results not reported in the original paper. SSL and LR stands for self-supervised learning and linear regression, respectively.}
  \label{tab:full_reference_results}
\end{table*}

We can easily extend our approach to the Full-Reference Image Quality Assessment (FR-IQA) task. FR-IQA aims to evaluate the quality of a distorted image in the setting in which a high-quality reference version is available. Similarly to \cite{madhusudana2022image, saha2023re}, we incorporate the information provided by the reference image with:
\begin{equation}
    y = W \left | h_{ref} - h_{dist} \right |
\end{equation}
where $y$ is the quality score, $W$ indicates the trainable weights of the regressor, and $h_{ref}$ and $h_{dist}$ are the representations of the reference and distorted image, respectively. Therefore, the regressor predicts the quality score associated with the difference between the embeddings of the reference and the distorted image.

We follow the same evaluation protocol described in Sec. {\color{red}4.3}, thereby not fine-tuning the encoder weights for the FR-IQA task. Note that we can only evaluate the performance on FR-IQA with datasets consisting of synthetic distortions, given the unavailability of a reference image for datasets with realistic degradations. We report the results in \cref{tab:full_reference_results}. Despite being designed for NR-IQA, \method obtains competitive results also on FR-IQA, thus further proving the effectiveness of our approach. Moreover, we observe that the additional information provided by the high-quality reference image leads to improved performance, compared to the NR-IQA setting reported in Tab. {\color{red}1}.

\subsection{Regressor Regularization Coefficient}
We recall that during evaluation we freeze the encoder weights and map the image representations to quality scores using simple linear regression, as in Re-IQA \cite{saha2023re}. Similarly to CONTRIQUE \cite{madhusudana2022image} and Re-IQA \cite{saha2023re}, we use the validation split of each dataset to identify the regularization coefficient of the Ridge regressor \cite{hoerl1970ridge} via a grid search over values within the range $\left[ 10^{-3}, 10^{3} \right]$. To assess the robustness of both \method and Re-IQA with respect to the choice of the regularization coefficient of the Ridge regression, we conduct an evaluation considering various values in the range $\left[ 10^{-3}, 10^{3} \right]$. \Cref{tab:regressor_regularization_coefficient} shows the results for the SRCC metric on the validation set of the \kadid dataset \cite{kadid10k}. As explained in Sec. {\color{red}4.3}, we report the median of the results of 10 random training/validation/test splits. We observe that our approach is significantly more robust than Re-IQA. In fact, the difference $\Delta$ between the best and worst results obtained for the various values of the regularization coefficient is considerably lower compared to Re-IQA.

\begin{table}
    \centering
    \begin{tabular}{lch}
        \toprule
        \multicolumn{1}{c}{} & \multicolumn{2}{c}{Method} \\
        \cmidrule(lr){2-3}
        Coefficient & Re-IQA$^{\dagger}$ & \textbf{\method} \\
        \midrule
        $\alpha \!=\! 0.001$ & 0.499 & 0.900 \\
        $\alpha \!=\! 0.01$ & 0.565 & 0.907 \\
        $\alpha \!=\! 0.1$ & 0.690 & 0.912 \\
        $\alpha \!=\! 1$ & 0.763 & 0.914 \\
        $\alpha \!=\! 10$ & 0.842 & 0.907 \\
        $\alpha \!=\! 100$ & 0.862 & 0.894 \\
        $\alpha \!=\! 1000$ & 0.858 & 0.859 \\ \midrule
        Best & 0.862 & \textbf{0.914} \\
        Worst & 0.499 & \textbf{0.859} \\
        $\Delta$ & 0.368 & \textbf{0.055} \\
        \bottomrule
    \end{tabular}
    \caption{Results for varying regressor regularization coefficient $\alpha$ for the SRCC metric on the validation set of the \kadid dataset \cite{kadid10k}. $\Delta$ indicates the difference between the best and worst scores. $^{\dagger}$ denotes results evaluated by us with the official pre-trained models. The best scores are highlighted in bold.}
    \label{tab:regressor_regularization_coefficient}
\end{table}

\subsection{gMAD Competition}
\begin{figure}
    \centering
    \begin{subfigure}{0.225\linewidth}
        \includegraphics[width=\linewidth]{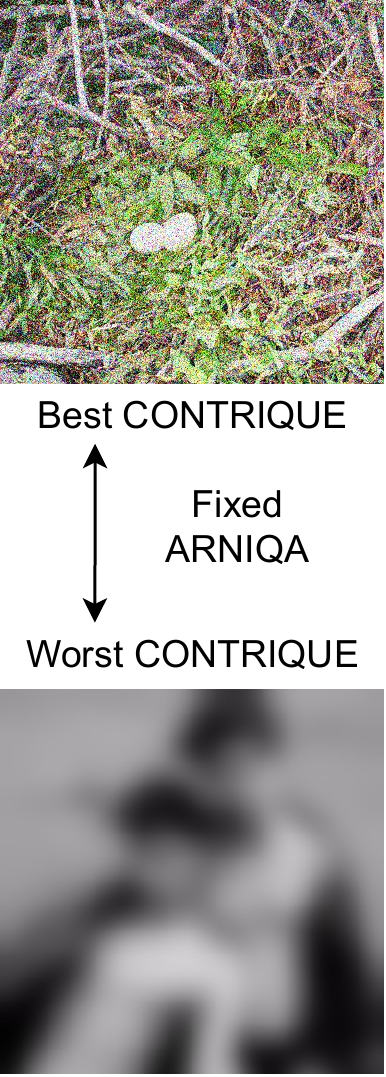}
        \caption{}
        \label{fig:ours_def_low}
    \end{subfigure}
    \hfill
    \begin{subfigure}{0.225\linewidth}
        \includegraphics[width=\linewidth]{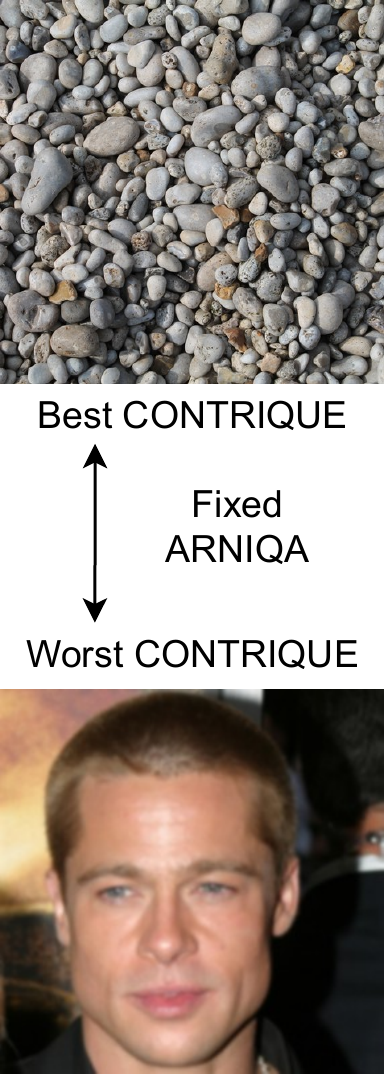}
        \caption{}
        \label{fig:ours_def_high}
    \end{subfigure}
    \hfill
    \begin{subfigure}{0.225\linewidth}
        \includegraphics[width=\linewidth]{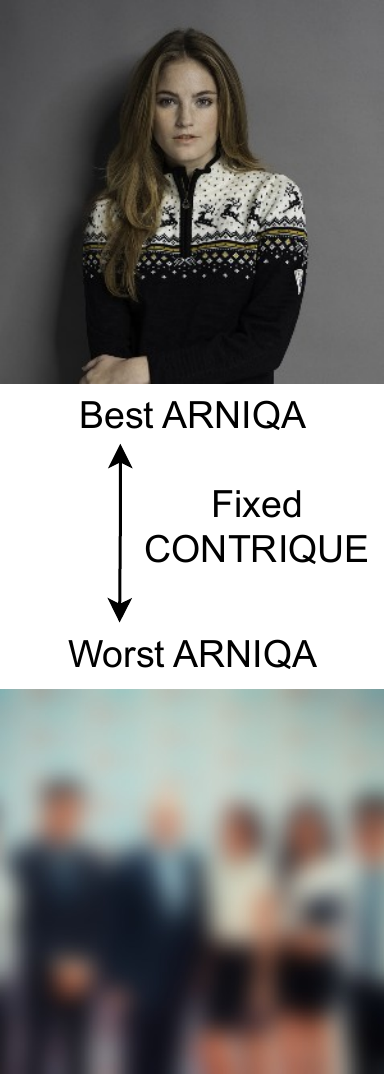}
        \caption{}
        \label{fig:ours_att_low}
    \end{subfigure}
    \hfill
    \begin{subfigure}{0.225\linewidth}
        \includegraphics[width=\linewidth]{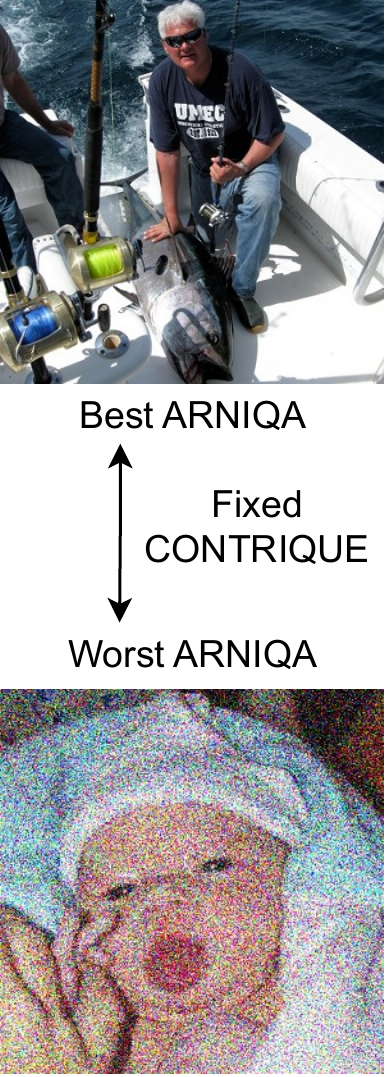}
        \caption{}
        \label{fig:ours_att_high}
    \end{subfigure}
    \caption{gMAD competition results between \method and CONTRIQUE \cite{madhusudana2022image}. (a) and (b): Fixed \method at a low- and high-quality level, respectively. (c) and (d): Fixed CONTRIQUE at a low- and high-quality level, respectively.}
    \label{fig:gmad_contrique}
\end{figure}

We conduct the group maximum differentiation (gMAD) competition \cite{ma2016group} between \method and CONTRIQUE \cite{madhusudana2022image} to evaluate the robustness of our model. See Sec. {\color{red}4.4} for more details about gMAD. We report the results in \cref{fig:gmad_contrique}. When we fix \method at a low-quality level (\cref{fig:ours_def_low}), CONTRIQUE struggles to identify picture pairs with a clear quality disparity. On the contrary, when fixing \method at a high-quality level, the image pair found by CONTRIQUE shows a slight divergence in quality. However, when acting as the attacker (\cref{fig:ours_att_low,fig:ours_att_high}), \method succeeds in highlighting the failures of CONTRIQUE by identifying image pairs exhibiting considerably different quality. Therefore, our method demonstrates superior robustness to that of CONTRIQUE. 

\subsection{Manifold Visualization}
We carry out an experiment to visualize the inherent structure of the distortion manifold learned by our model. Given two distortion types, our aim is to study the positions occupied in the manifold by images that exhibit both single and combined degradation patterns with varying levels of intensity. For a model that effectively learned the image distortion manifold, we expect images showing combined degradation patterns to occupy positions within the manifold that are intermediate to the locations associated with the single distortions themselves.

To conduct this study, we consider 1000 randomly selected pristine images from the KADIS dataset \cite{kadid10k} and the Gaussian blur and white noise distortions (see \cref{sec:distortion_types} for more details). First, we distort the images individually with each of the two degradations under consideration, using 5 different levels of intensity. Then,
we consider all the possible combinations of the degrees of intensity of the Gaussian blur and white noise distortions, taken in this order. Finally, we distort each of the pristine images with each combination by applying the two distortions consecutively. Therefore, for each image, we obtain 5 + 5 embeddings corresponding to the single blur and noise distortions, and 5 $\times$ 5 representations for the combined ones.

\begin{figure}
    \centering
    \begin{subfigure}{0.9\linewidth}
        \captionsetup{skip=8pt}
        \includegraphics[width=\linewidth]{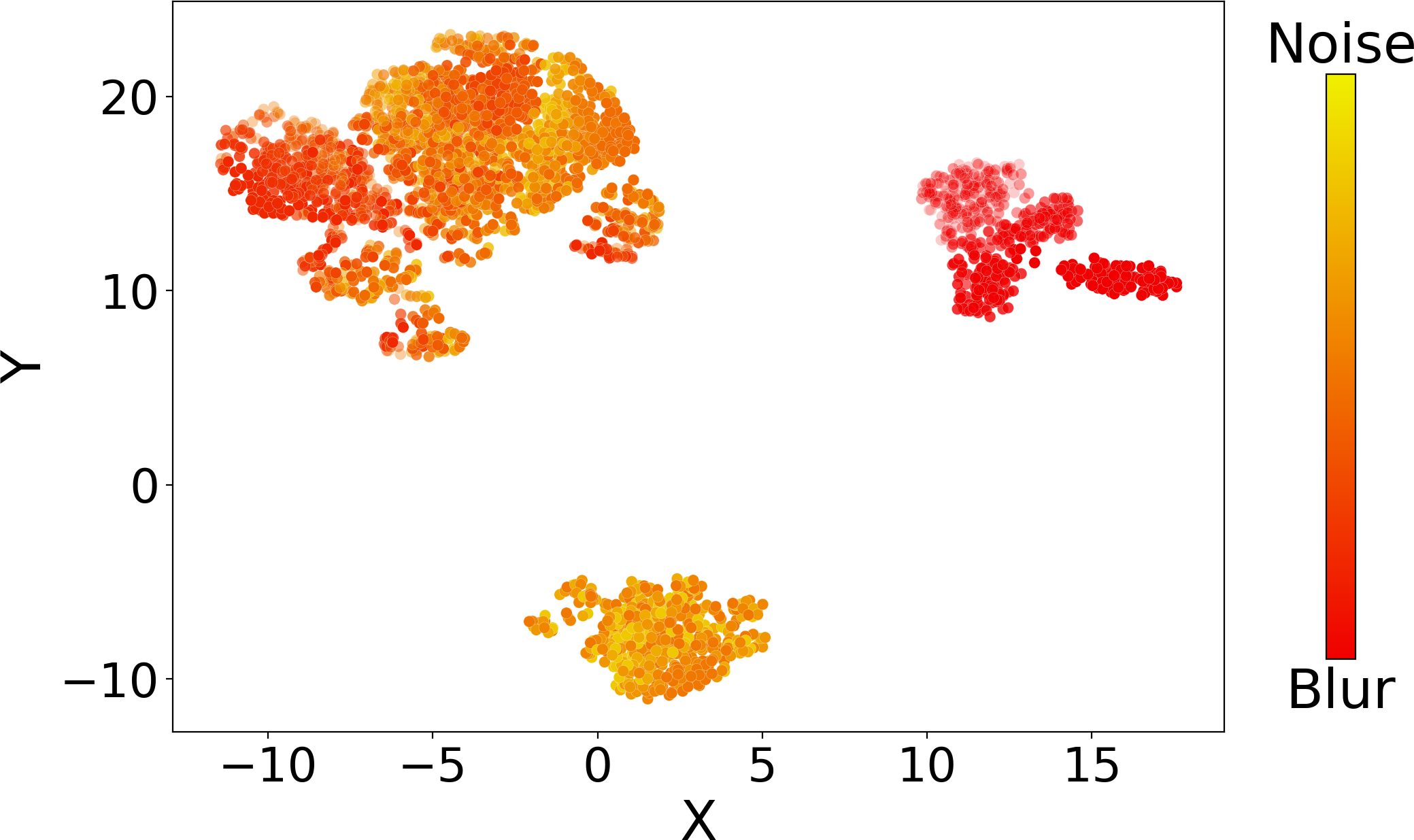}
        \caption{Re-IQA \cite{saha2023re} \vspace{3ex}}
        \label{fig:reiqa_manifold_visualization}
    \end{subfigure}
    \begin{subfigure}{0.9\linewidth}
        \captionsetup{skip=8pt}
        \includegraphics[width=\linewidth]{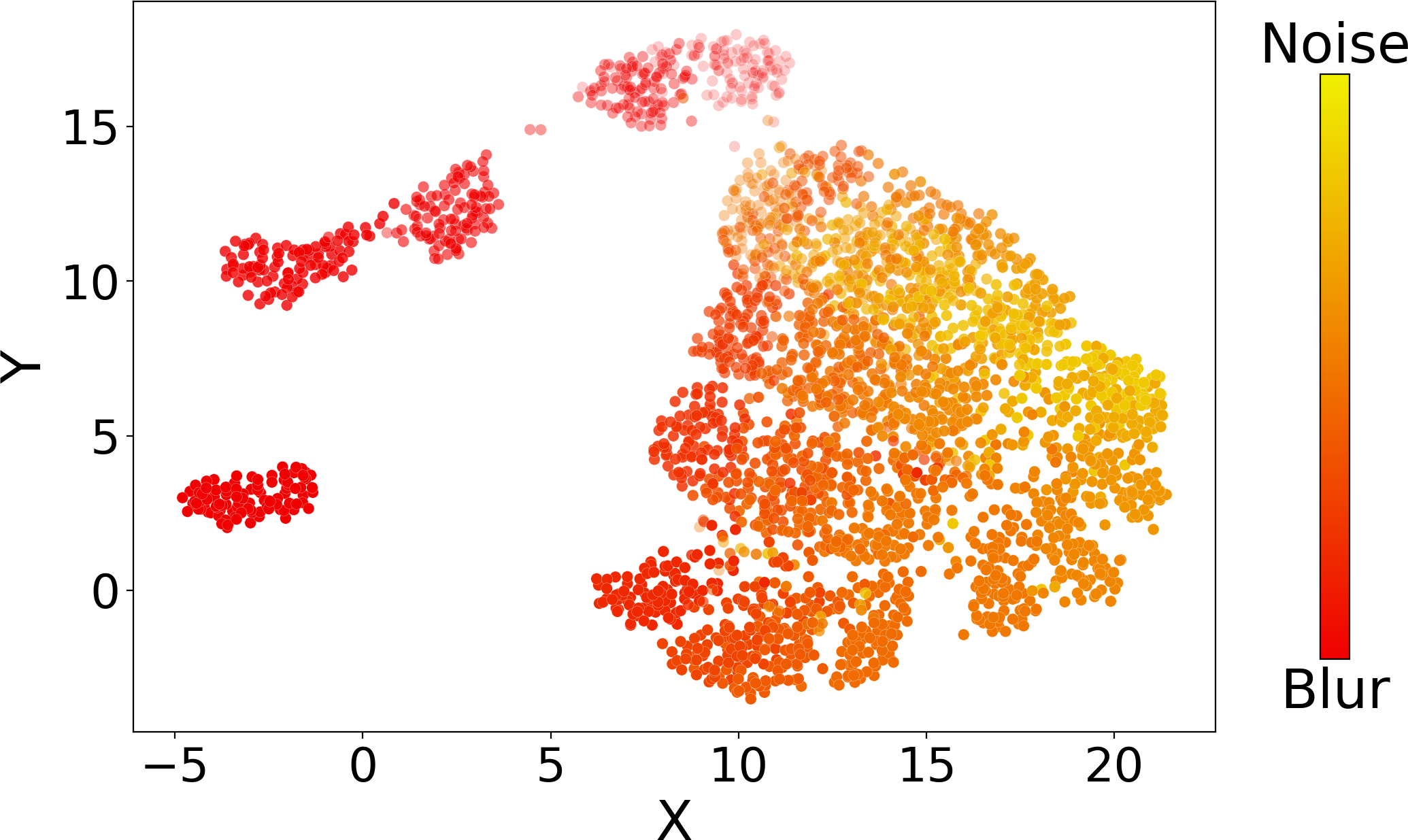}
        \caption{\method}
        \label{fig:ours_manifold_visualization}
    \end{subfigure}
    
    \caption{Manifold visualization with UMAP \cite{mcinnes2018umap} of the embeddings of 1000 images degraded with Gaussian blur and white noise distortions, applied in this order. The color of each point is given by the weighted average between the colors of blur (red) and noise (yellow), based on the degradation intensity. A higher alpha value corresponds to a stronger degradation intensity.}
    \label{fig:manifold_visualization}
\end{figure}

\Cref{fig:reiqa_manifold_visualization,fig:ours_manifold_visualization} shows the UMAP visualization \cite{mcinnes2018umap} of the embeddings obtained with Re-IQA \cite{saha2023re} and \method, respectively. As we can see, compared to Re-IQA, our approach leads to a smoother transition between the points corresponding to the single and combined degradations. Indeed, the stronger the intensity of the noise distortion, the closer the points are to the cluster of images degraded only with white noise. Note that most of the points corresponding to combined degradation patterns lie closer to the cluster of images distorted only with white noise as it was applied after the blur. Indeed, the final degradation in a distortion composition corresponds to more visible patterns, as they are not modified by subsequent degradations.

\section{Image Degradation Model}

\subsection{Distortion Compositions}
\begin{figure*}
    \centering
    \setlength{\tabcolsep}{1pt}
    \renewcommand{\arraystretch}{0.5}
    \resizebox{\textwidth}{!}{
    \begin{tabular}{cccccccccc}
        \includegraphics{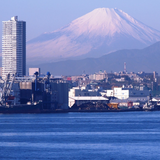} & \includegraphics{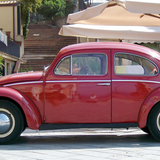} & \includegraphics{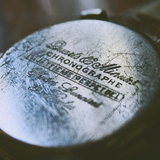} & \includegraphics{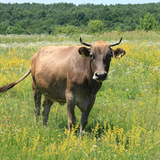} & \includegraphics{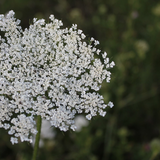} & \includegraphics{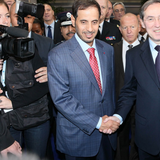} & \includegraphics{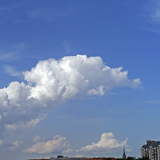} & \includegraphics{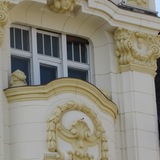} & \includegraphics{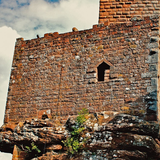} & \includegraphics{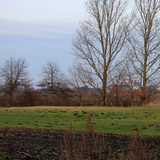} \\
        \includegraphics{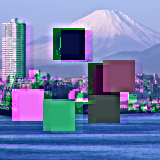} & \includegraphics{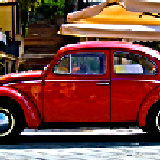} & \includegraphics{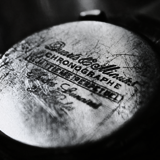} & \includegraphics{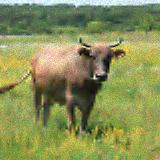} & \includegraphics{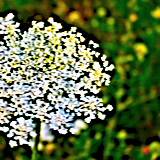} & \includegraphics{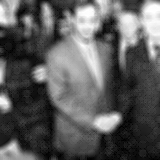} & \includegraphics{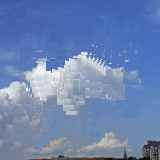} & \includegraphics{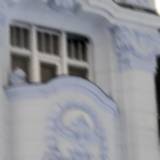}  & \includegraphics{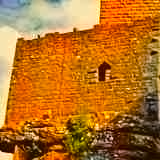} & \includegraphics{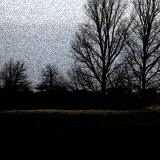} \\
    \end{tabular}
    }
    \caption{Comparison between pristine images from the KADIS dataset \cite{kadid10k} and their distorted versions using the proposed degradation model. \textit{Top}: Pristine images. \textit{Bottom}: Distorted images.}
    \label{fig:distortion_compositions}
\end{figure*}

In \cref{fig:distortion_compositions} we report some examples of images belonging to the KADIS dataset \cite{kadid10k} subjected to distortion compositions obtained through our image degradation model. We notice how the proposed degradation model leads to images showing a large variety of distortion patterns. In this way, our model is able to effectively learn the image distortion manifold.

\subsection{Distortion Types} \label{sec:distortion_types}
\begin{figure*}
    \centering
    \setlength{\tabcolsep}{1pt}
    \Large
    \resizebox{\textwidth}{!}{ 
\begin{tabular}{C{6em}ccccc}
         & Level 1 & Level 2 & Level 3 & Level 4 & Level 5 \\
 Brighten & \includegraphics[width=\gridimagewidth,valign=m]{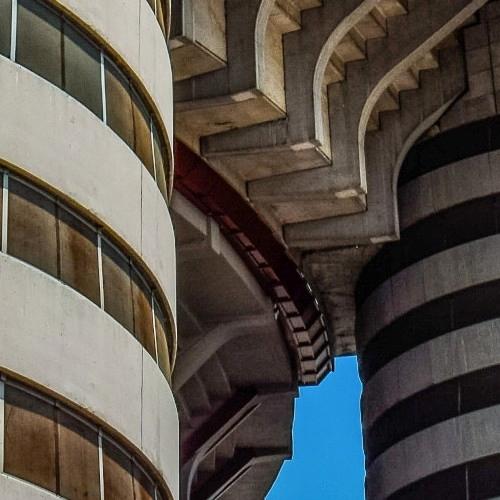} & \includegraphics[width=\gridimagewidth,valign=m]{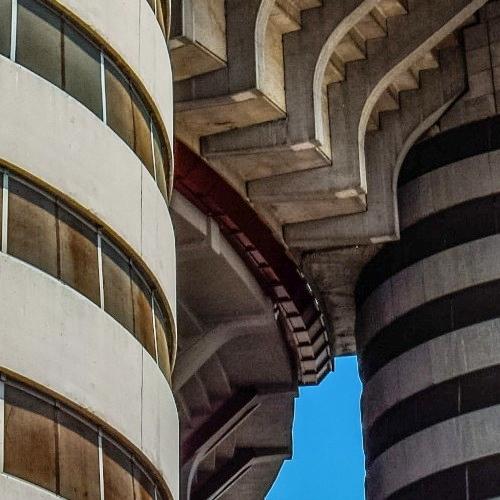} & \includegraphics[width=\gridimagewidth,valign=m]{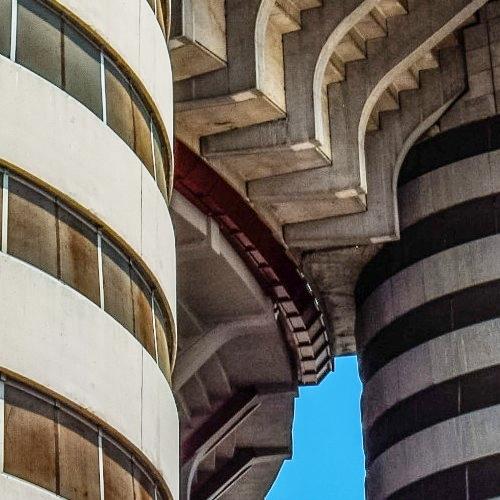} & \includegraphics[width=\gridimagewidth,valign=m]{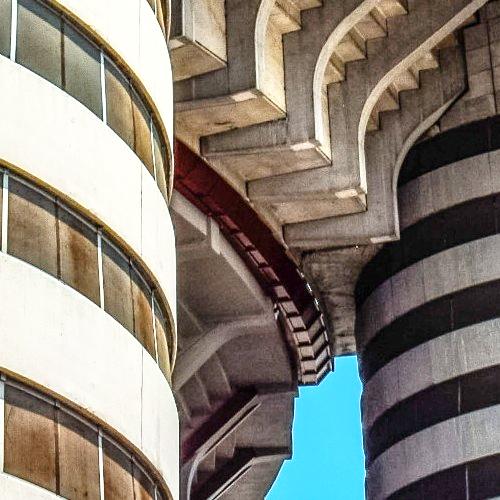} & \includegraphics[width=\gridimagewidth,valign=m]{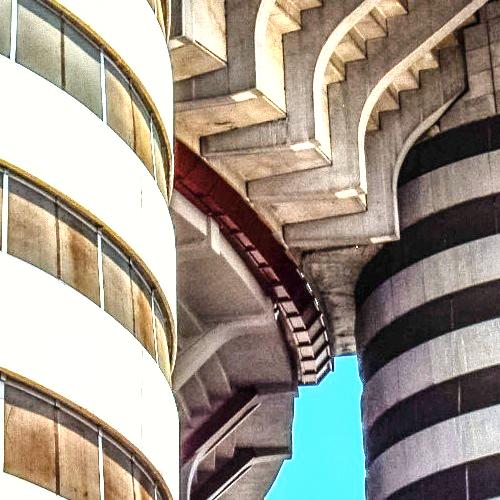} \\ [6.15ex]
 Darken & \includegraphics[width=\gridimagewidth,valign=m]{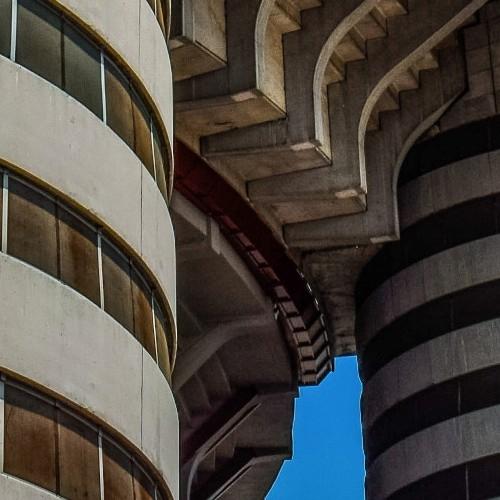} & \includegraphics[width=\gridimagewidth,valign=m]{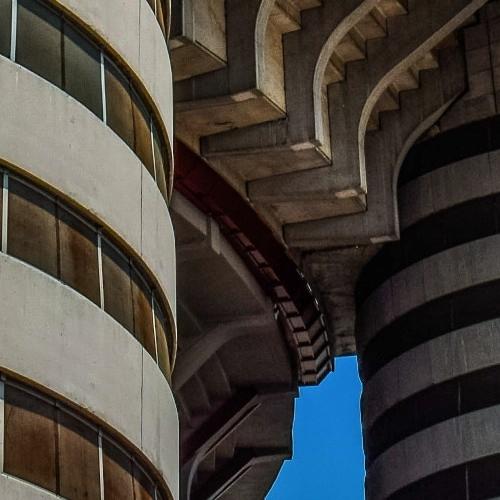} & \includegraphics[width=\gridimagewidth,valign=m]{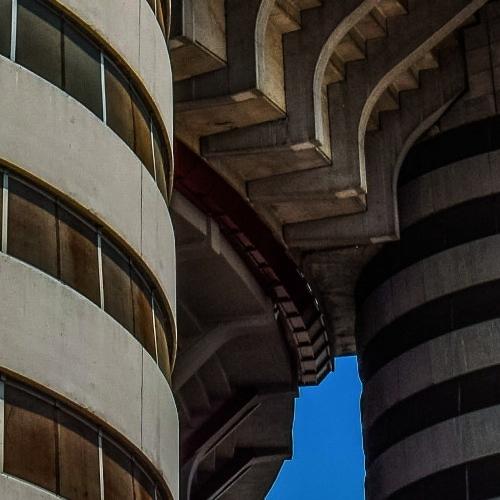} & \includegraphics[width=\gridimagewidth,valign=m]{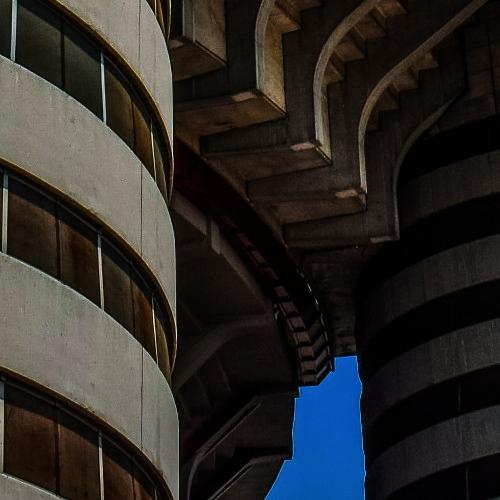} & \includegraphics[width=\gridimagewidth,valign=m]{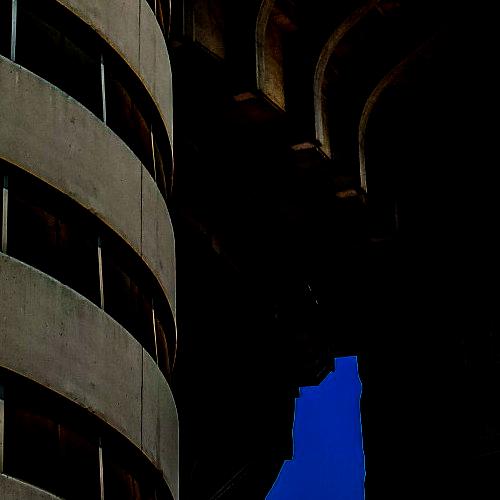} \\ [6.15ex]
 Mean shift & \includegraphics[width=\gridimagewidth,valign=m]{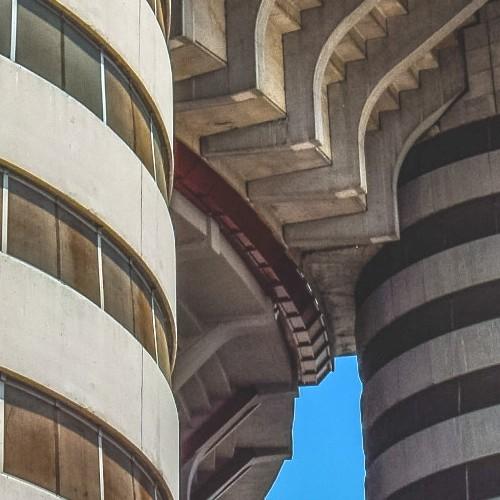} & 
 \includegraphics[width=\gridimagewidth,valign=m]{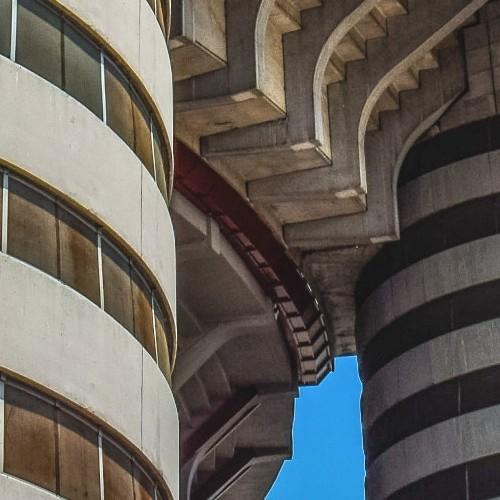} & \includegraphics[width=\gridimagewidth,valign=m]{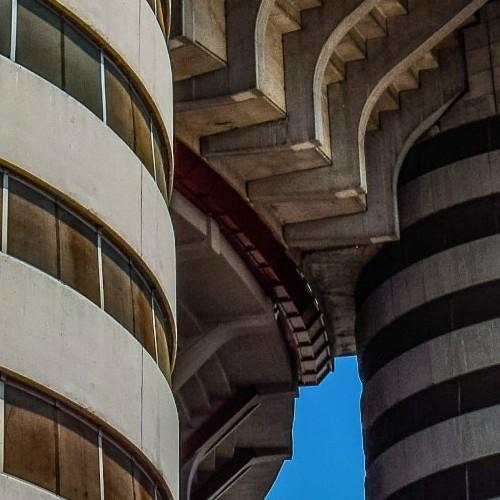} & \includegraphics[width=\gridimagewidth,valign=m]{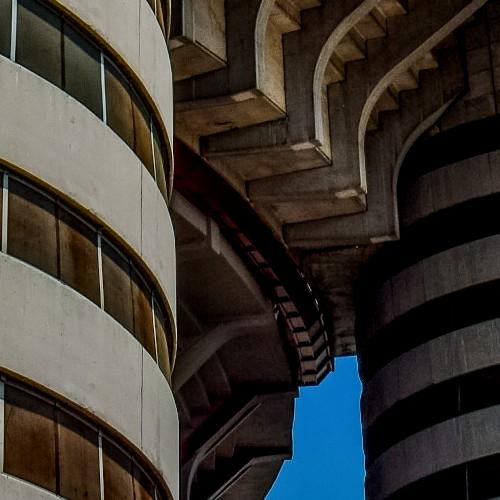} & \includegraphics[width=\gridimagewidth,valign=m]{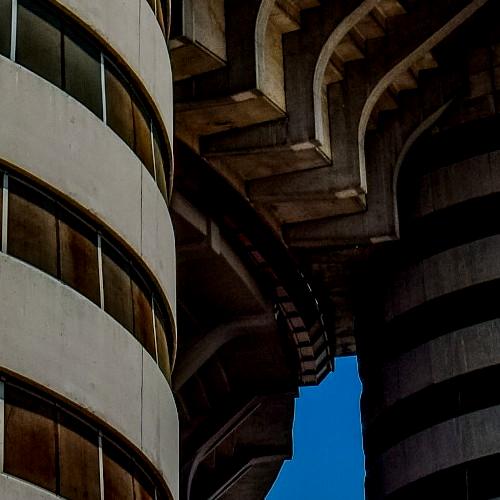} \\
 
\end{tabular}
}
\caption{Visualization of the degradation types belonging to the \textit{Brightness change} group for increasing levels of intensity.}
\label{fig:brightness_change}
\end{figure*}

\begin{figure*}
    \centering
    \setlength{\tabcolsep}{1pt}
    \Large
    \resizebox{\textwidth}{!}{ 
\begin{tabular}{C{6em}ccccc}
         & Level 1 & Level 2 & Level 3 & Level 4 & Level 5 \\
 Gaussian blur & \includegraphics[width=\gridimagewidth,valign=m]{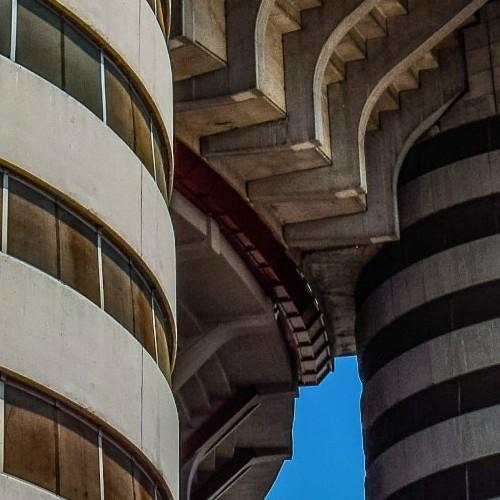} & \includegraphics[width=\gridimagewidth,valign=m]{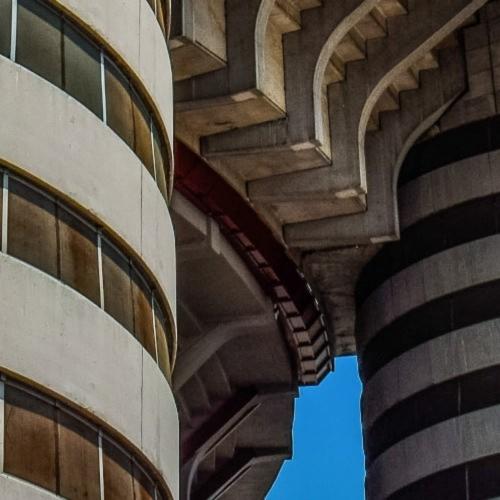} & \includegraphics[width=\gridimagewidth,valign=m]{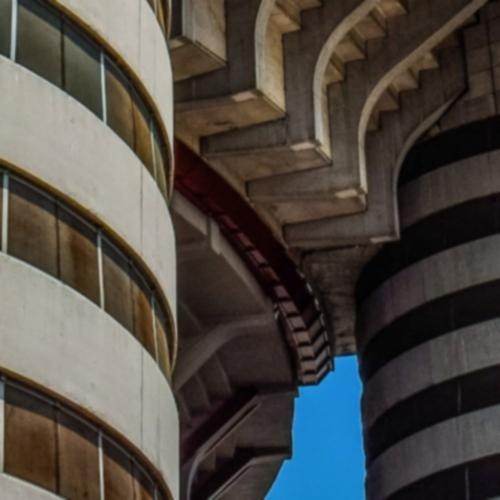} & \includegraphics[width=\gridimagewidth,valign=m]{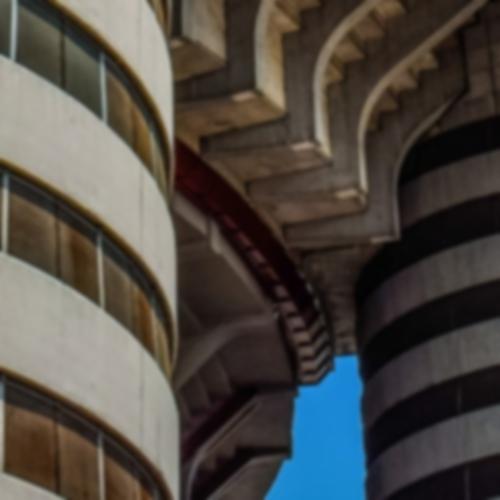} & \includegraphics[width=\gridimagewidth,valign=m]{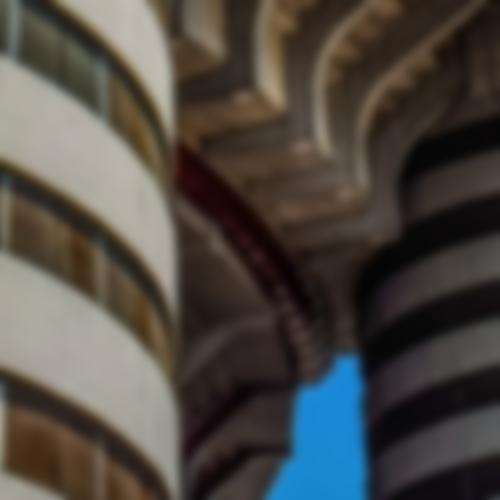} \\ [6.15ex]
 Lens blur & \includegraphics[width=\gridimagewidth,valign=m]{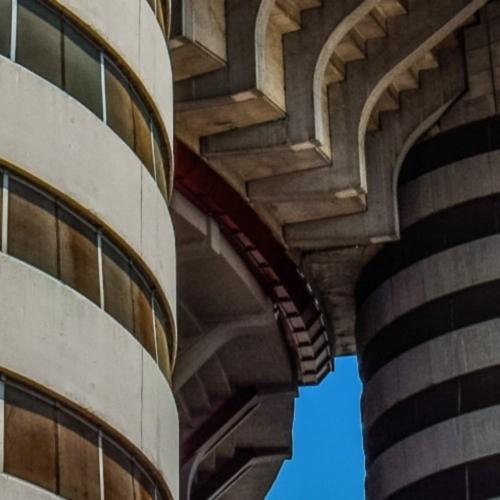} & \includegraphics[width=\gridimagewidth,valign=m]{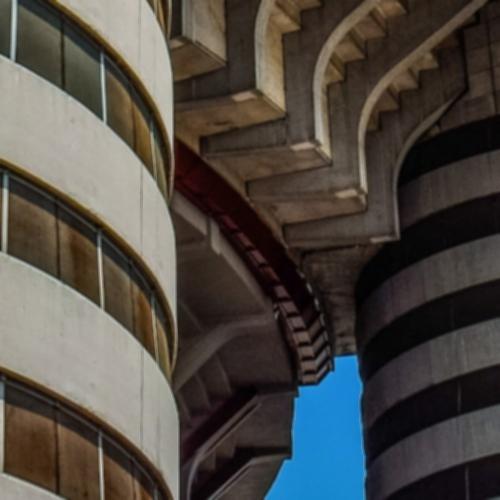} & \includegraphics[width=\gridimagewidth,valign=m]{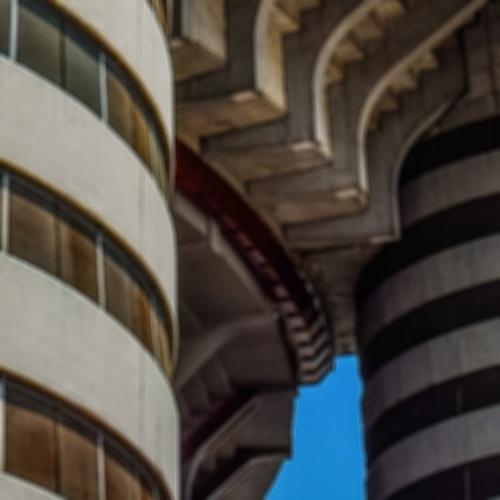} & \includegraphics[width=\gridimagewidth,valign=m]{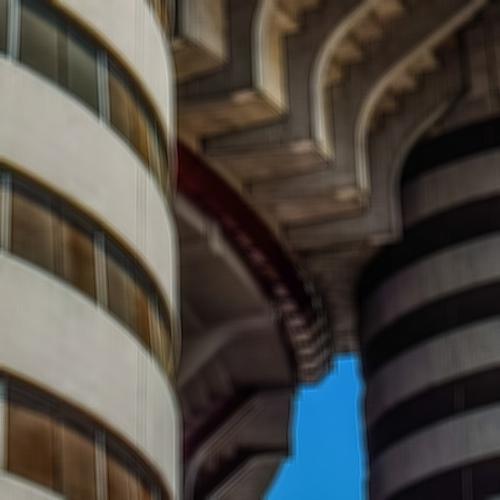} & \includegraphics[width=\gridimagewidth,valign=m]{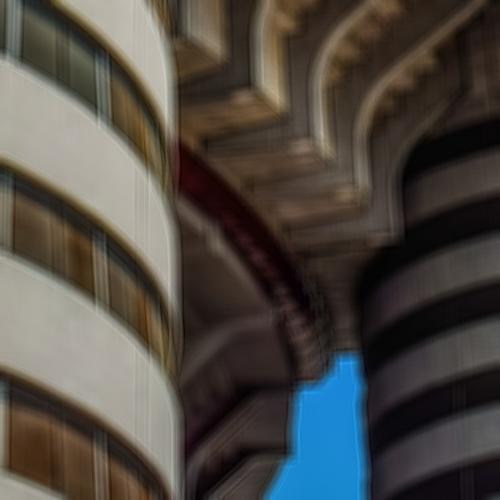} \\ [6.15ex]
 Motion blur & \includegraphics[width=\gridimagewidth,valign=m]{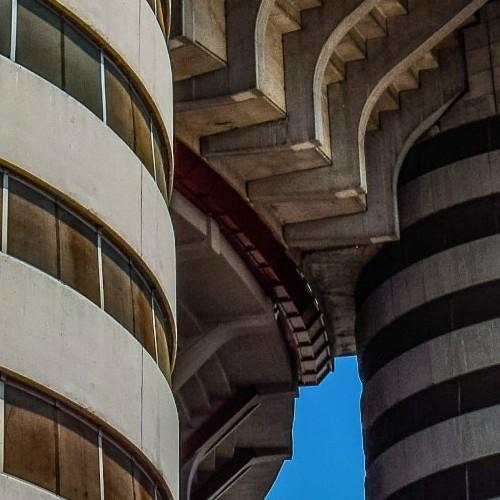} & \includegraphics[width=\gridimagewidth,valign=m]{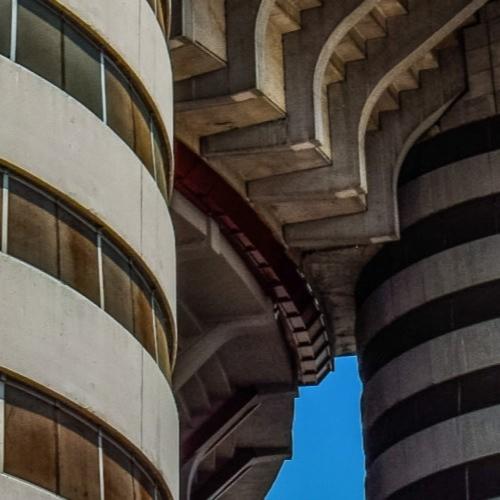} & \includegraphics[width=\gridimagewidth,valign=m]{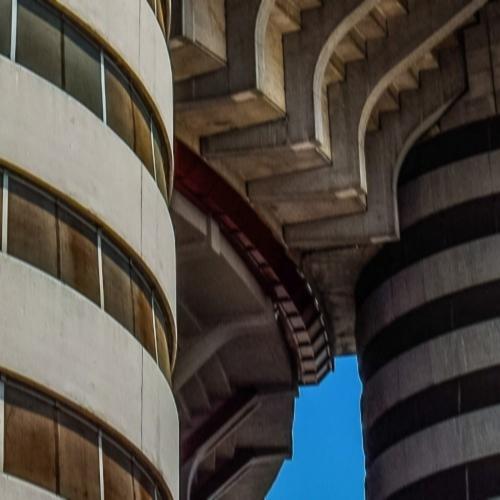} & \includegraphics[width=\gridimagewidth,valign=m]{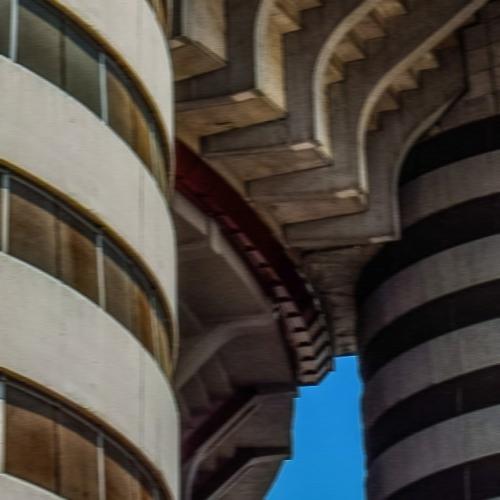} & \includegraphics[width=\gridimagewidth,valign=m]{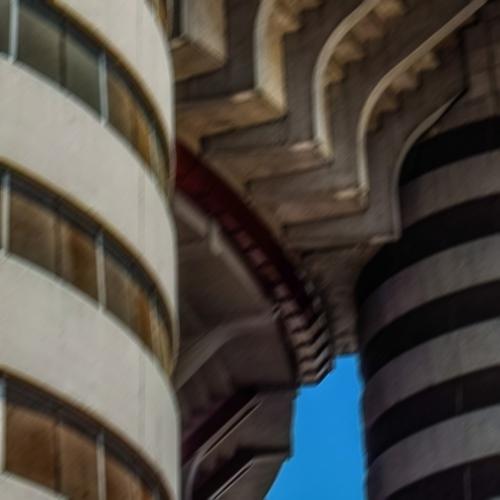} \\ 
\end{tabular}
}
\caption{Visualization of the degradation types belonging to the \textit{Blur} group for increasing levels of intensity.}
\label{fig:blur}
\end{figure*}

\begin{figure*}
    \centering
    \setlength{\tabcolsep}{1pt}
    \Large
    \resizebox{\textwidth}{!}{ 
\begin{tabular}{C{6em}ccccc}
         & Level 1 & Level 2 & Level 3 & Level 4 & Level 5 \\
 Jitter & \includegraphics[width=\gridimagewidth,valign=m]{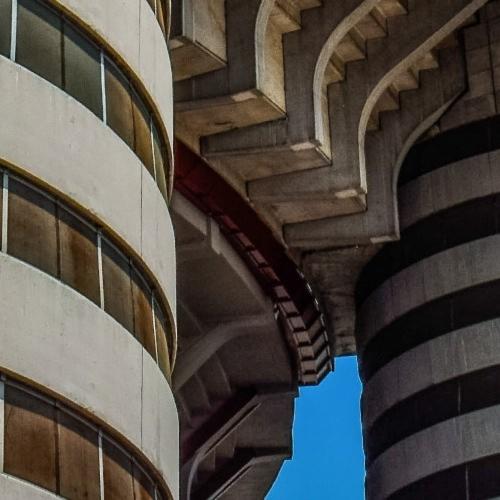} & \includegraphics[width=\gridimagewidth,valign=m]{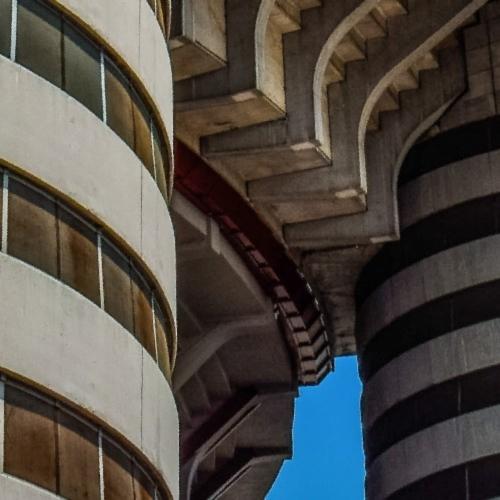} & \includegraphics[width=\gridimagewidth,valign=m]{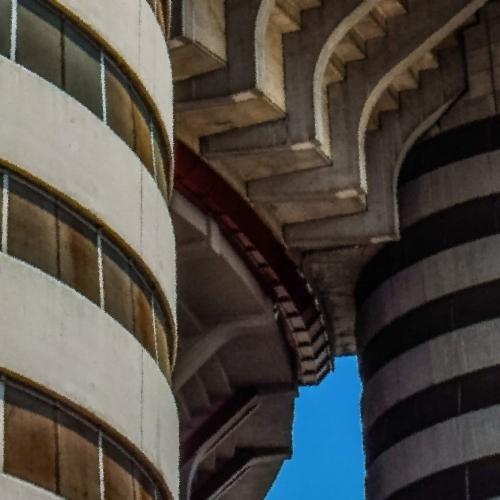} & \includegraphics[width=\gridimagewidth,valign=m]{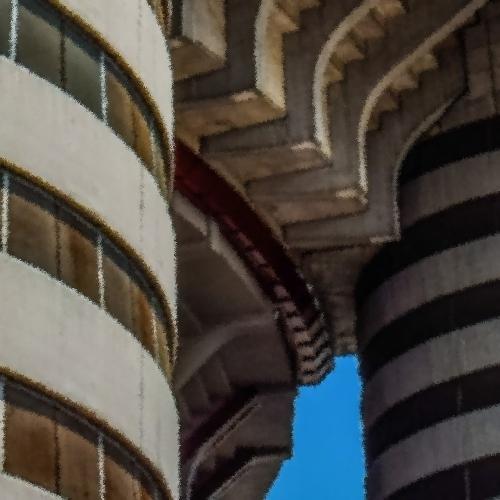} & \includegraphics[width=\gridimagewidth,valign=m]{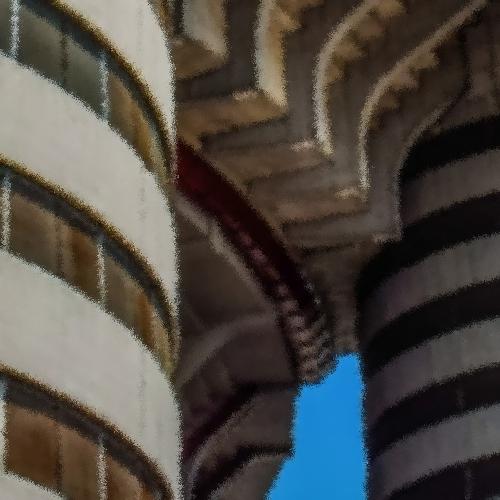} \\ [6.15ex]
 Non-eccentricity patch & \includegraphics[width=\gridimagewidth,valign=m]{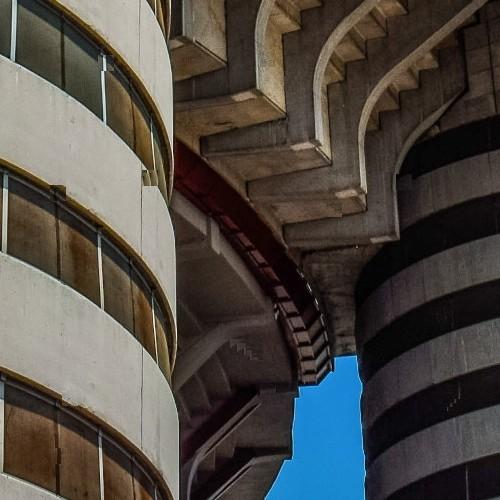} & \includegraphics[width=\gridimagewidth,valign=m]{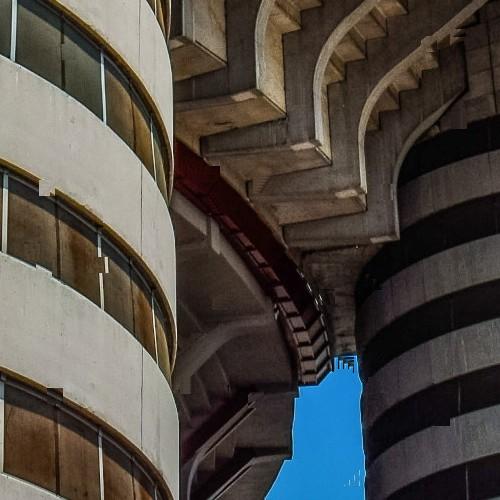} & \includegraphics[width=\gridimagewidth,valign=m]{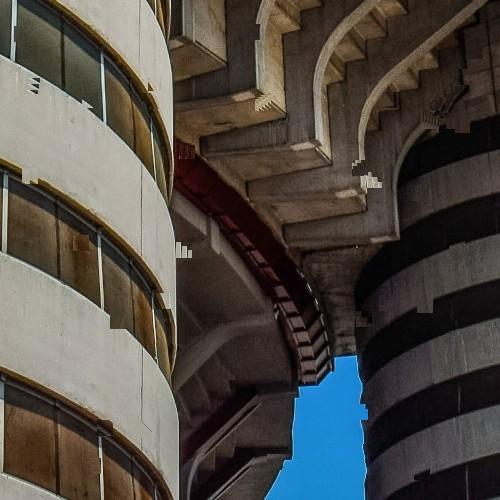} & \includegraphics[width=\gridimagewidth,valign=m]{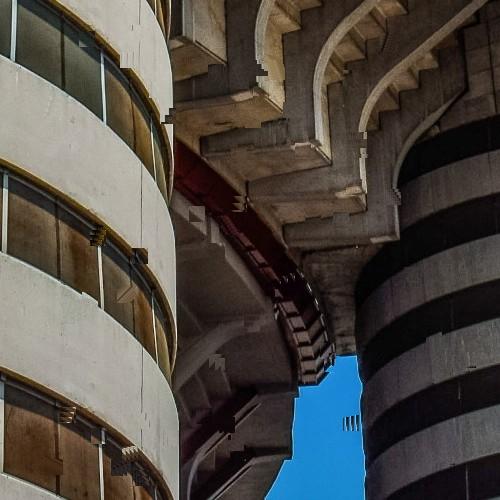} & \includegraphics[width=\gridimagewidth,valign=m]{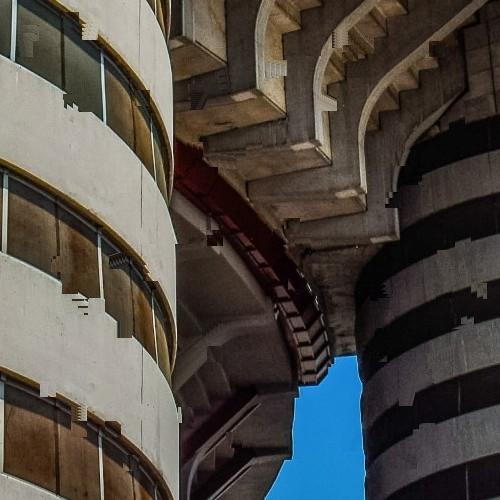} \\ [6.15ex]
 Pixelate & \includegraphics[width=\gridimagewidth,valign=m]{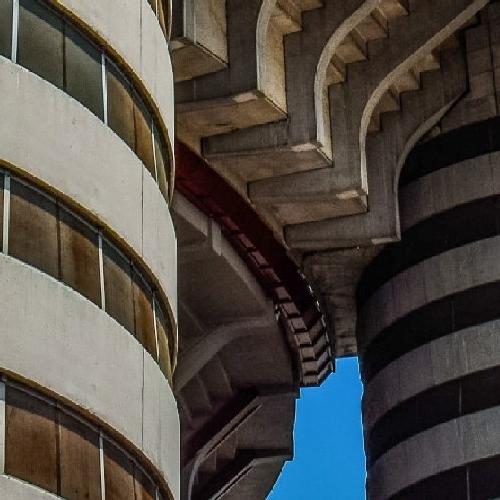} & \includegraphics[width=\gridimagewidth,valign=m]{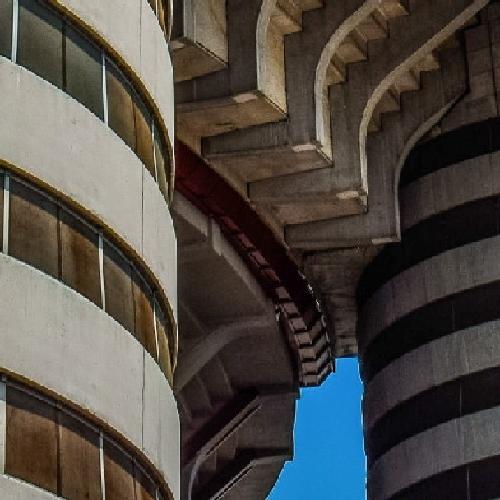} & \includegraphics[width=\gridimagewidth,valign=m]{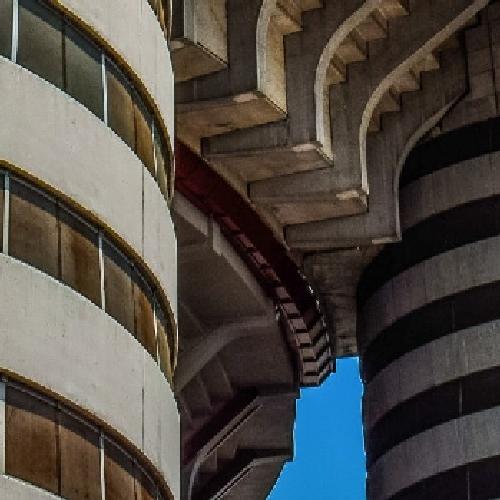} & \includegraphics[width=\gridimagewidth,valign=m]{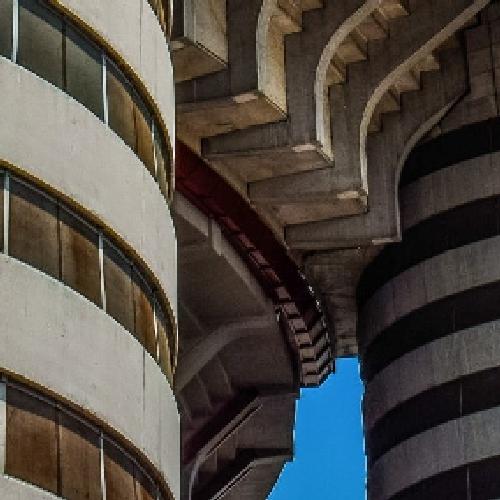} & \includegraphics[width=\gridimagewidth,valign=m]{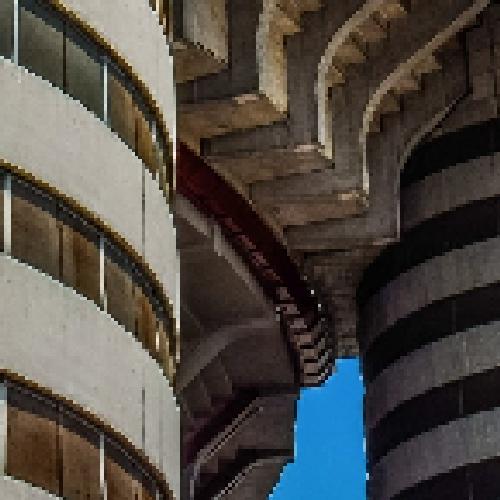} \\ [6.15ex]
 Quantization & \includegraphics[width=\gridimagewidth,valign=m]{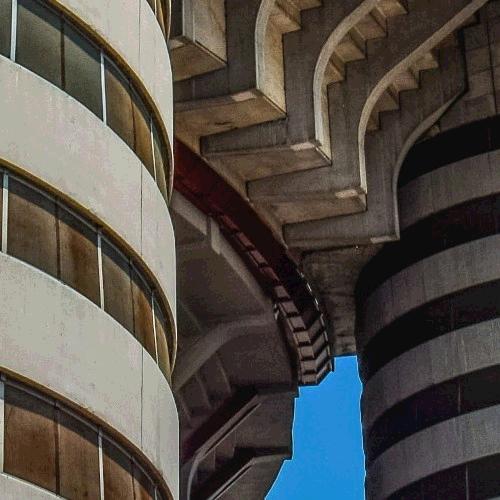} & \includegraphics[width=\gridimagewidth,valign=m]{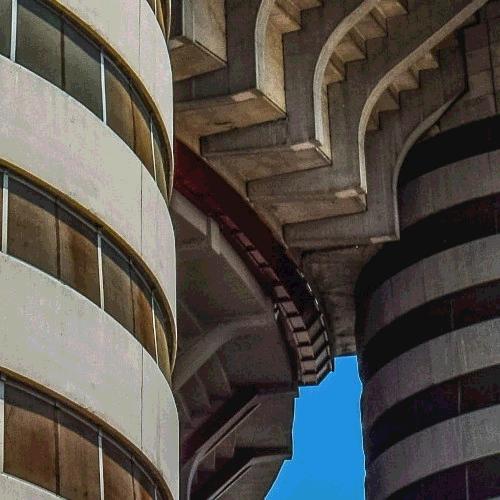} & \includegraphics[width=\gridimagewidth,valign=m]{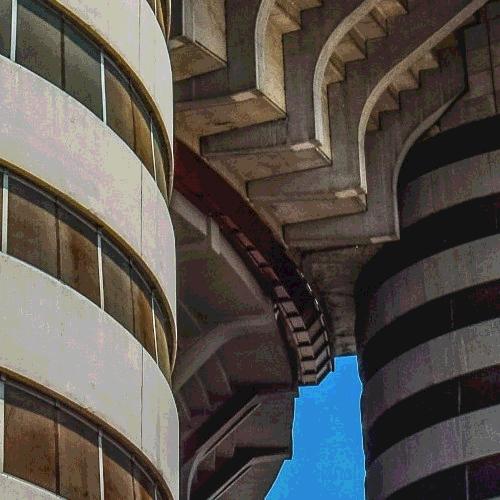} & \includegraphics[width=\gridimagewidth,valign=m]{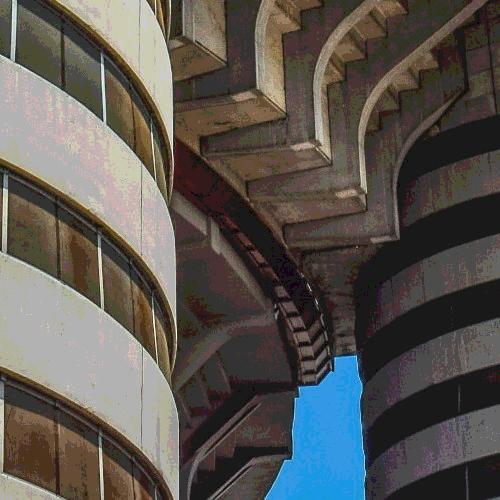} & \includegraphics[width=\gridimagewidth,valign=m]{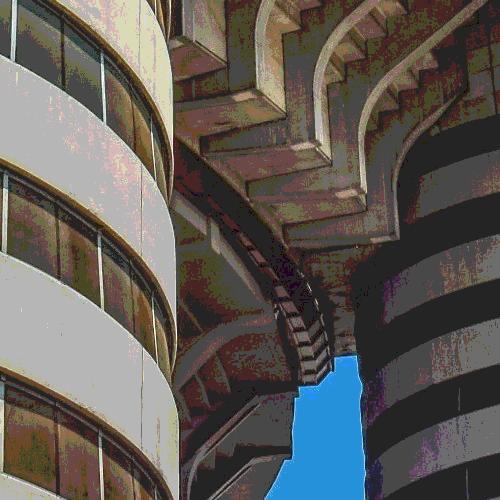} \\ [6.15ex]
 Color block & \includegraphics[width=\gridimagewidth,valign=m]{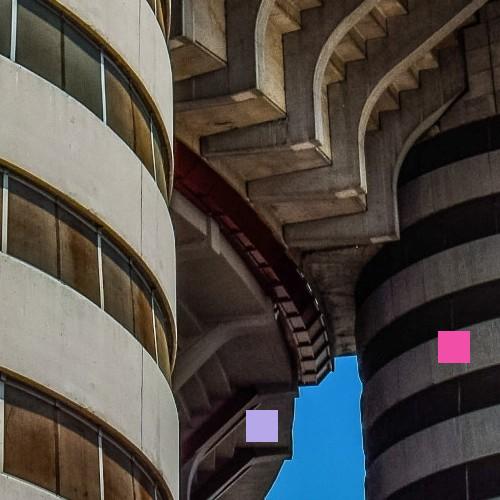} & \includegraphics[width=\gridimagewidth,valign=m]{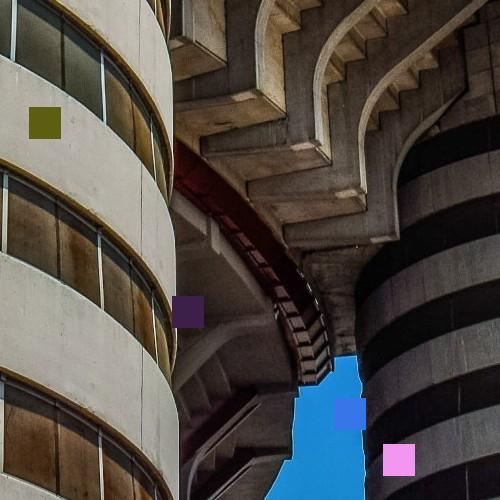} & \includegraphics[width=\gridimagewidth,valign=m]{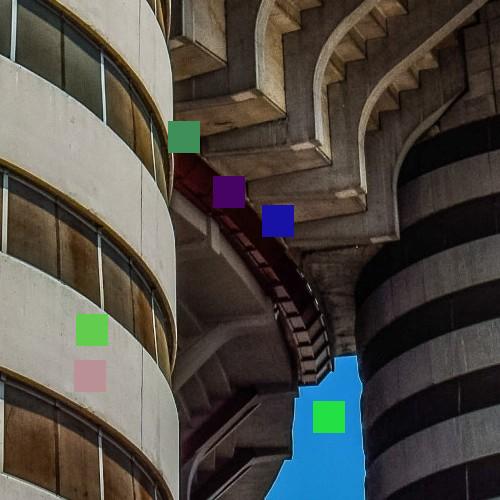} & \includegraphics[width=\gridimagewidth,valign=m]{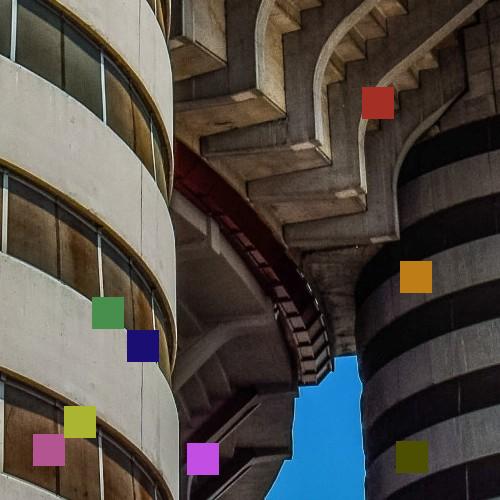} & \includegraphics[width=\gridimagewidth,valign=m]{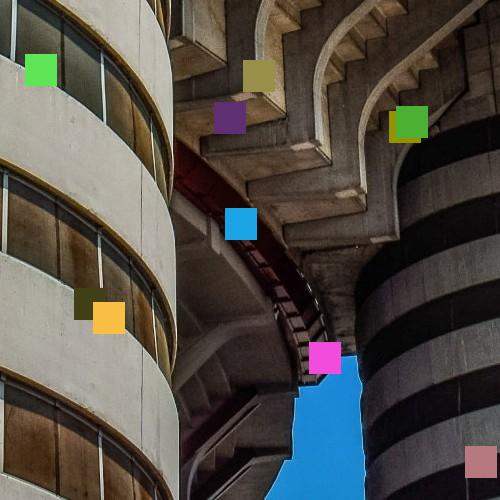} \\
\end{tabular}
}
\caption{Visualization of the degradation types belonging to the \textit{Spatial distortions} group for increasing levels of intensity.}
\label{fig:spatial_distortions}
\end{figure*}

\begin{figure*}
    \centering
    \setlength{\tabcolsep}{1pt}
    \Large
    \resizebox{\textwidth}{!}{ 
\begin{tabular}{C{6em}ccccc}
         & Level 1 & Level 2 & Level 3 & Level 4 & Level 5 \\
 White noise & \includegraphics[width=\gridimagewidth,valign=m]{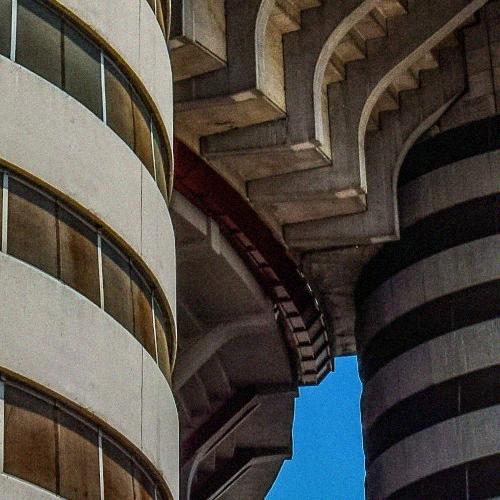} & \includegraphics[width=\gridimagewidth,valign=m]{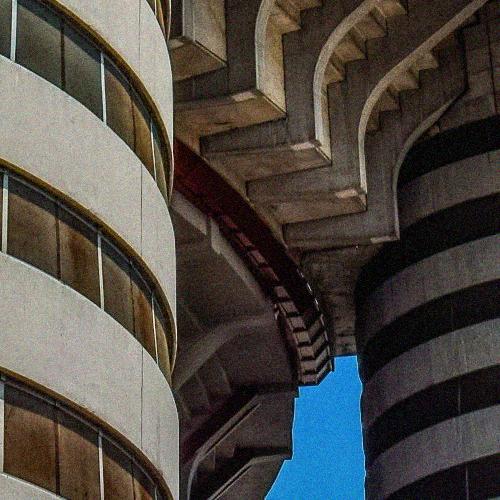} & \includegraphics[width=\gridimagewidth,valign=m]{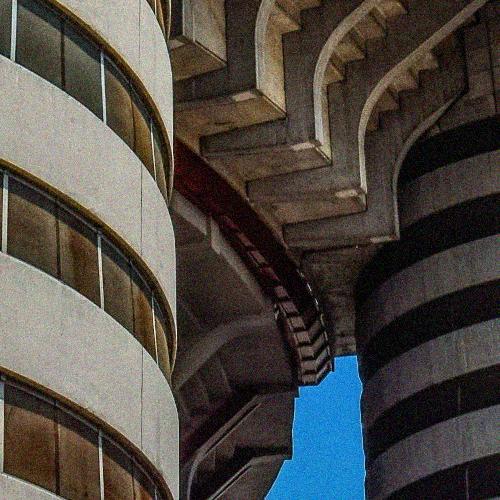} & \includegraphics[width=\gridimagewidth,valign=m]{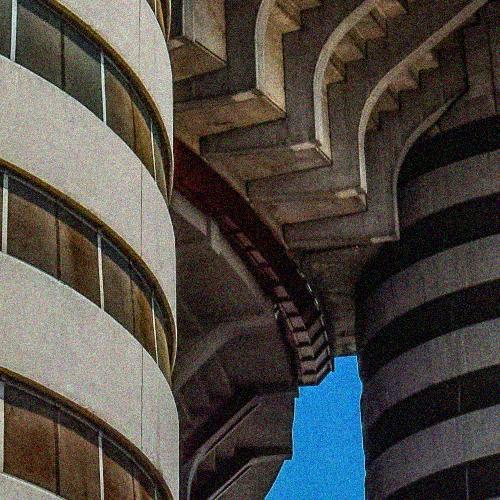} & \includegraphics[width=\gridimagewidth,valign=m]{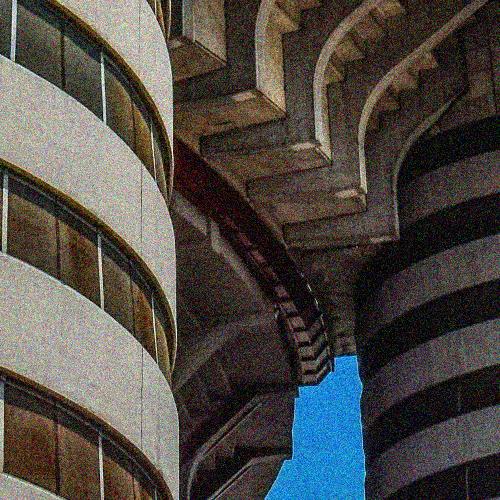} \\ [6.15ex]
 White noise \newline color \newline component & \includegraphics[width=\gridimagewidth,valign=m]{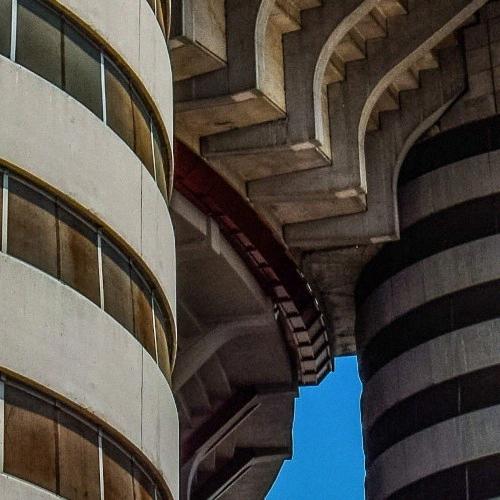} & \includegraphics[width=\gridimagewidth,valign=m]{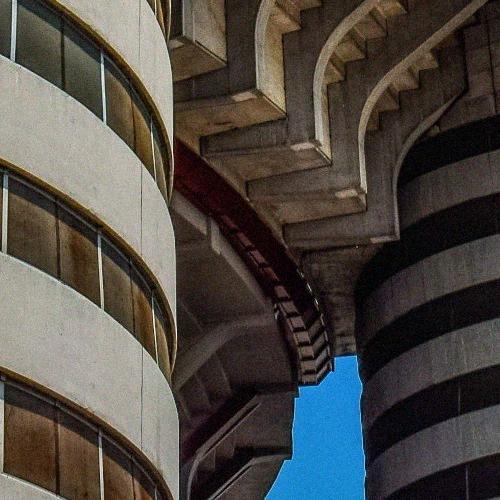} & \includegraphics[width=\gridimagewidth,valign=m]{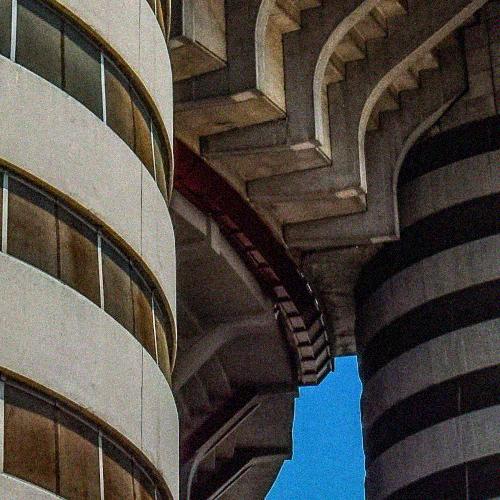} & \includegraphics[width=\gridimagewidth,valign=m]{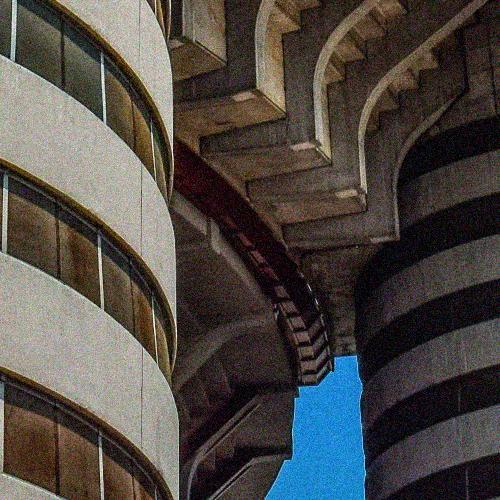} & \includegraphics[width=\gridimagewidth,valign=m]{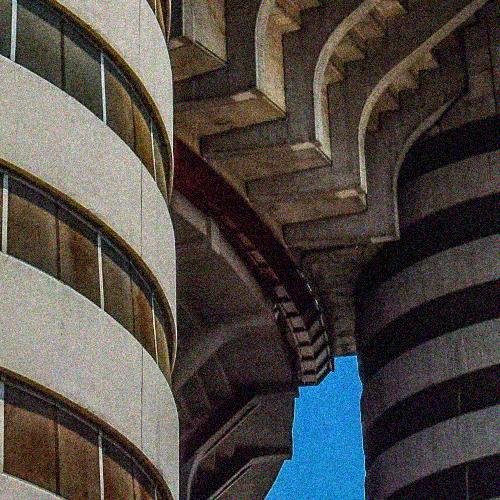} \\ [6.15ex]
 Impulse noise & \includegraphics[width=\gridimagewidth,valign=m]{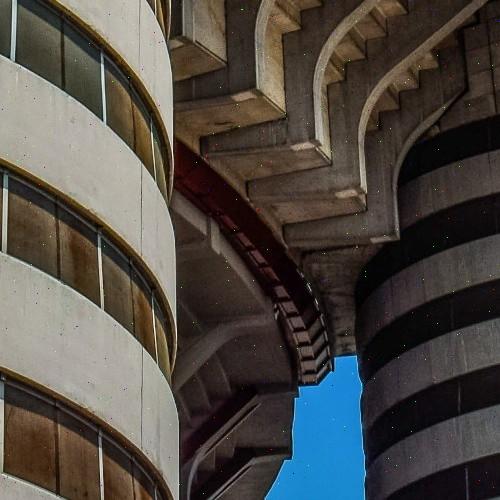} & \includegraphics[width=\gridimagewidth,valign=m]{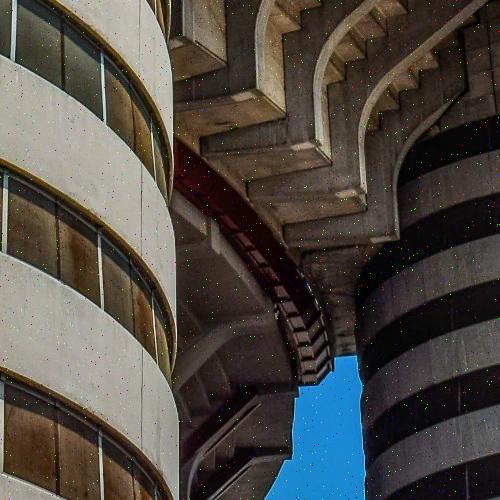} & \includegraphics[width=\gridimagewidth,valign=m]{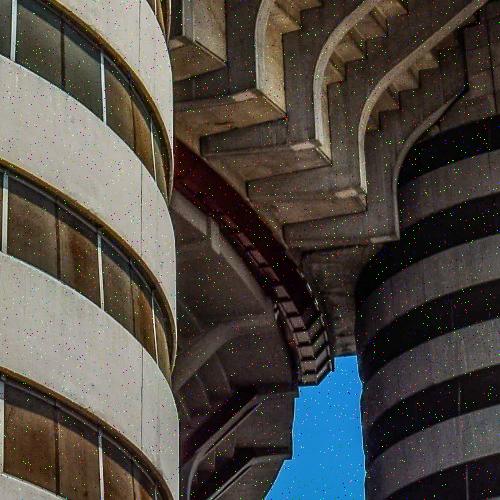} & \includegraphics[width=\gridimagewidth,valign=m]{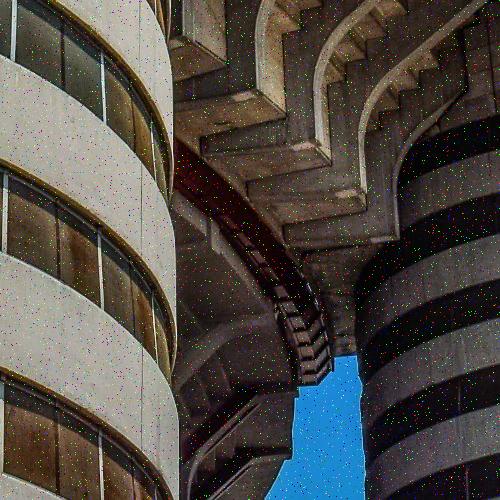} & \includegraphics[width=\gridimagewidth,valign=m]{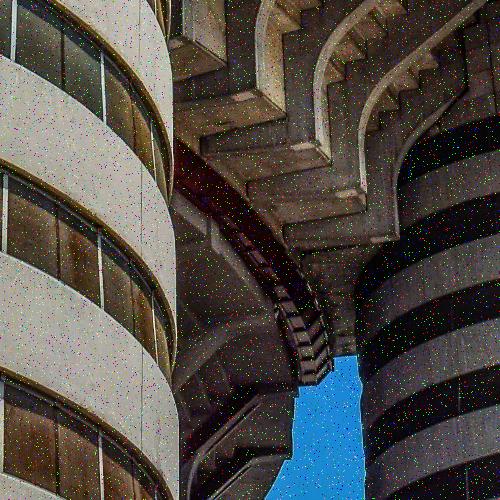} \\ [6.15ex]
 Multiplicative noise & \includegraphics[width=\gridimagewidth,valign=m]{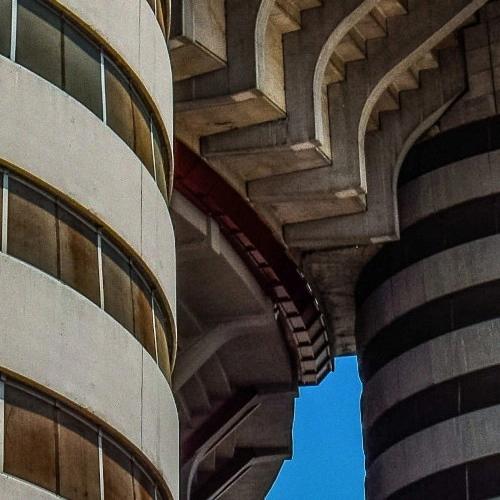} & \includegraphics[width=\gridimagewidth,valign=m]{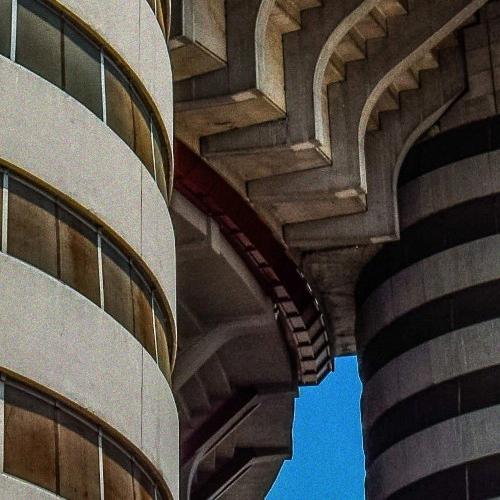} & \includegraphics[width=\gridimagewidth,valign=m]{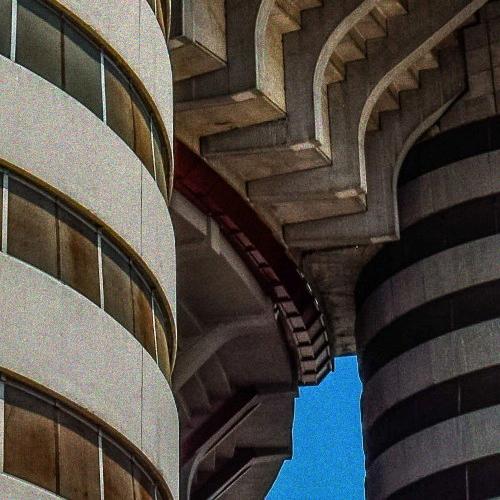} & \includegraphics[width=\gridimagewidth,valign=m]{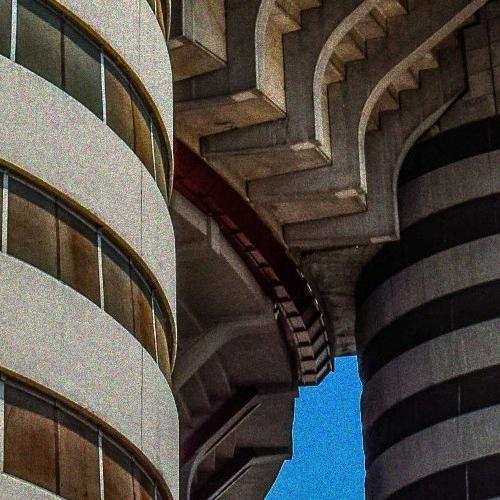} & \includegraphics[width=\gridimagewidth,valign=m]{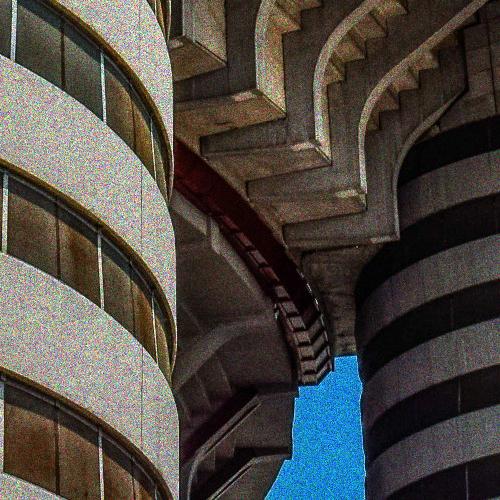} \\
\end{tabular}
}
\caption{Visualization of the degradation types belonging to the \textit{Noise} group for increasing levels of intensity.}
\label{fig:noise}
\end{figure*}

\begin{figure*}
    \centering
    \setlength{\tabcolsep}{1pt}
    \Large
    \resizebox{\textwidth}{!}{ 
\begin{tabular}{C{6em}ccccc}
         & Level 1 & Level 2 & Level 3 & Level 4 & Level 5 \\
 Color diffusion & \includegraphics[width=\gridimagewidth,valign=m]{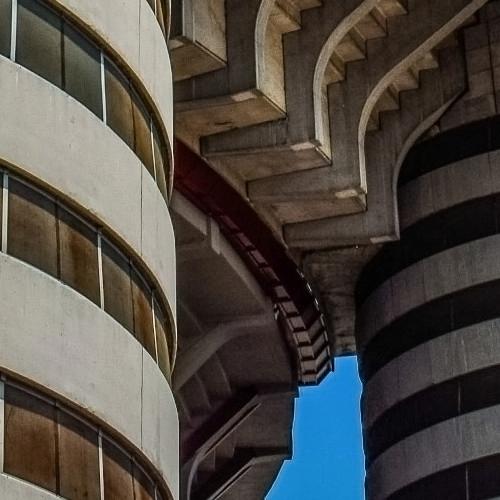} & \includegraphics[width=\gridimagewidth,valign=m]{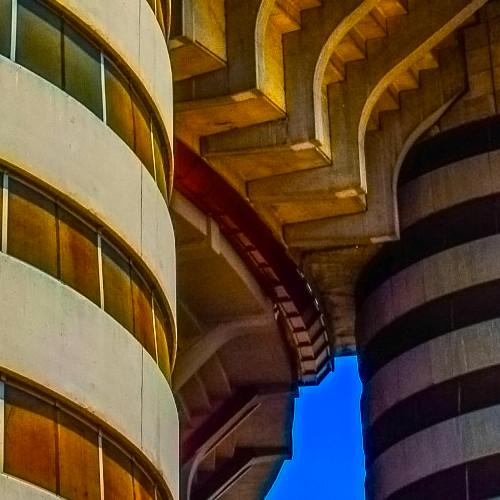} & \includegraphics[width=\gridimagewidth,valign=m]{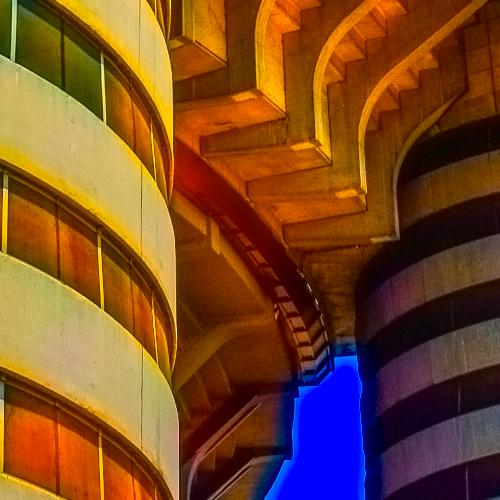} & \includegraphics[width=\gridimagewidth,valign=m]{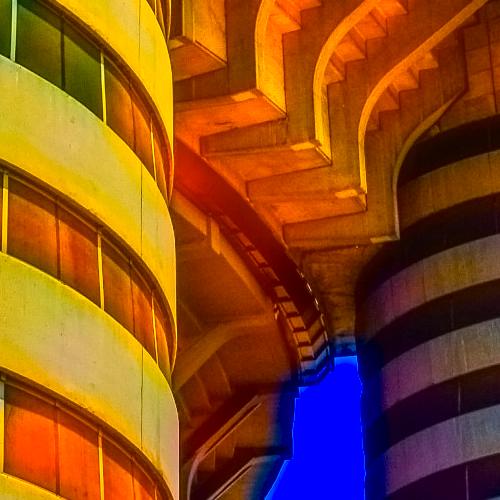} & \includegraphics[width=\gridimagewidth,valign=m]{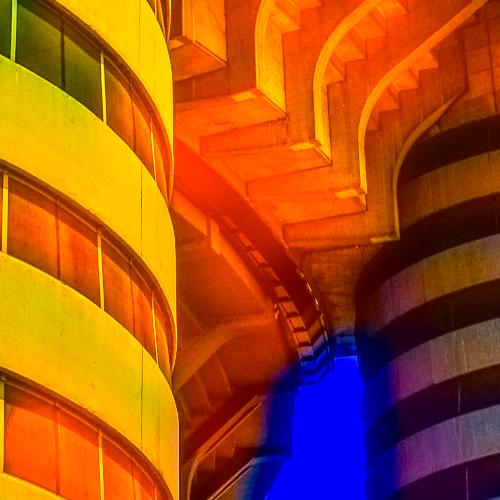} \\ [6.15ex]
 Color shift & \includegraphics[width=\gridimagewidth,valign=m]{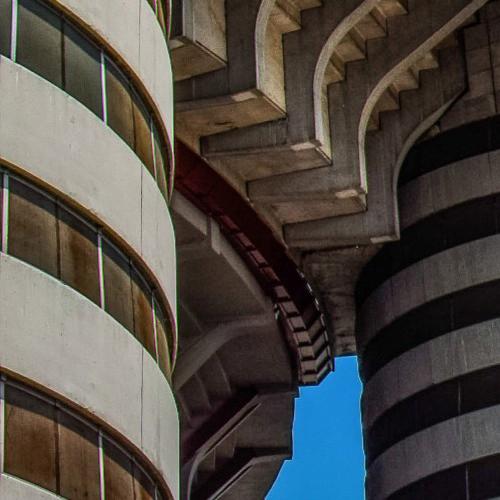} & \includegraphics[width=\gridimagewidth,valign=m]{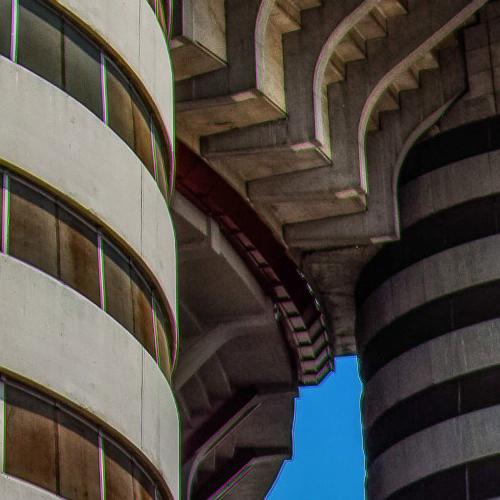} & \includegraphics[width=\gridimagewidth,valign=m]{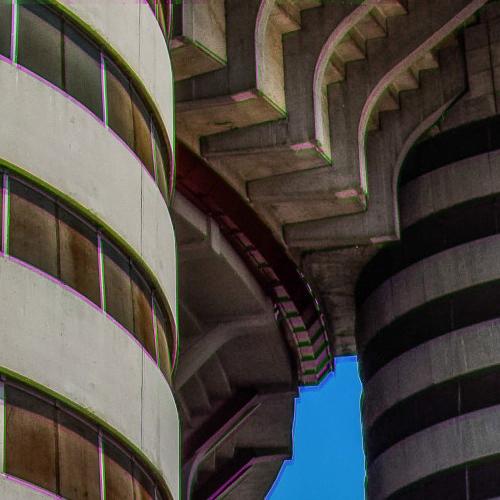} & \includegraphics[width=\gridimagewidth,valign=m]{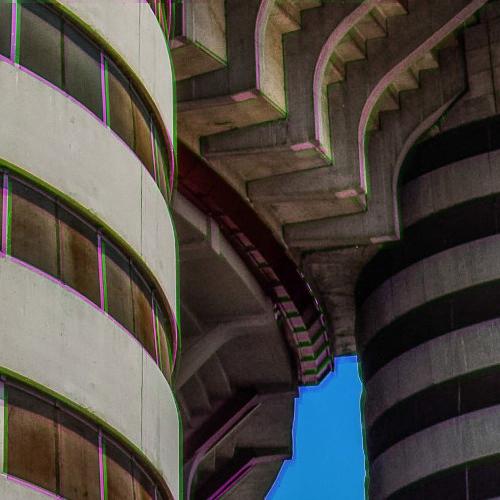} & \includegraphics[width=\gridimagewidth,valign=m]{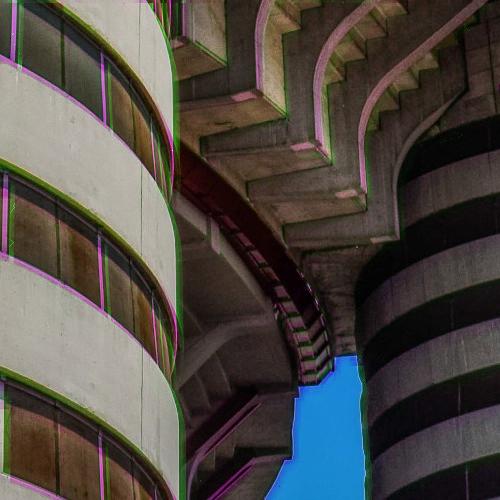} \\ [6.15ex]
 Color saturation 1 & \includegraphics[width=\gridimagewidth,valign=m]{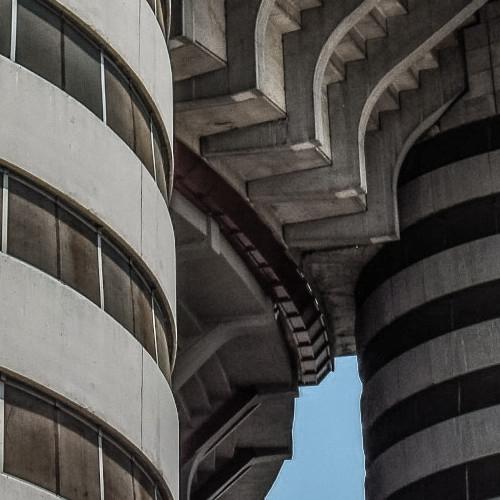} & \includegraphics[width=\gridimagewidth,valign=m]{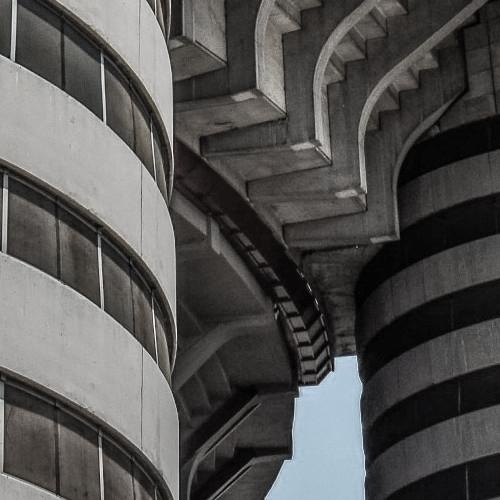} & \includegraphics[width=\gridimagewidth,valign=m]{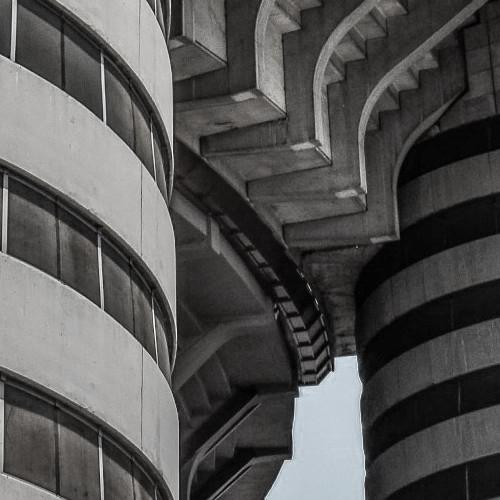} & \includegraphics[width=\gridimagewidth,valign=m]{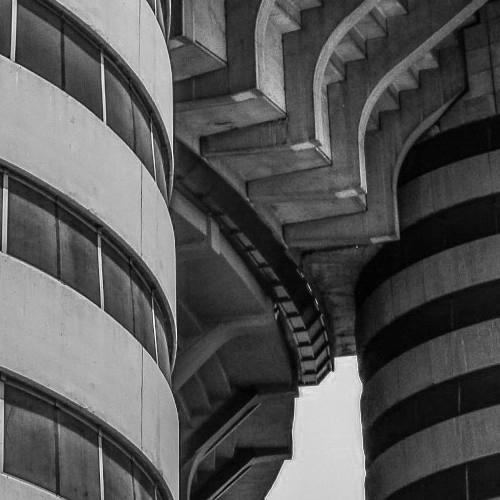} & \includegraphics[width=\gridimagewidth,valign=m]{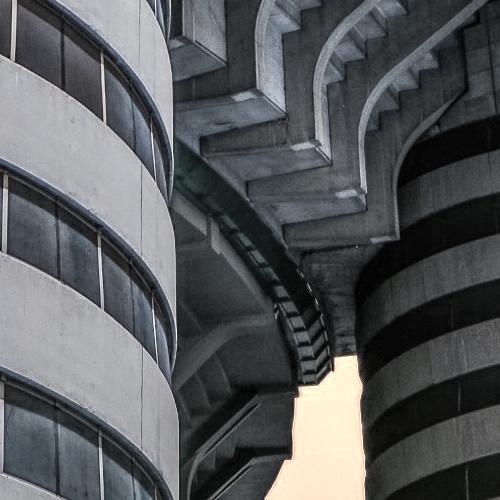} \\ [6.15ex]
 Color saturation 2 & \includegraphics[width=\gridimagewidth,valign=m]{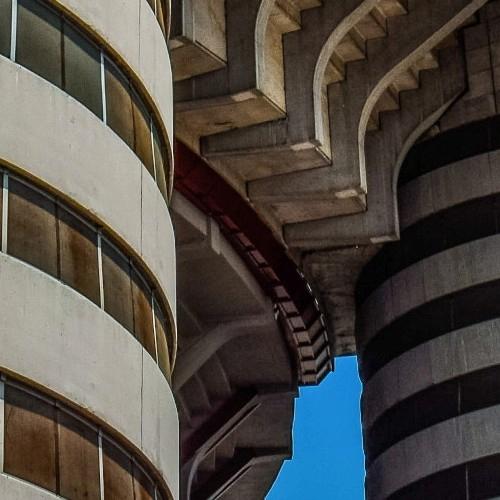} & \includegraphics[width=\gridimagewidth,valign=m]{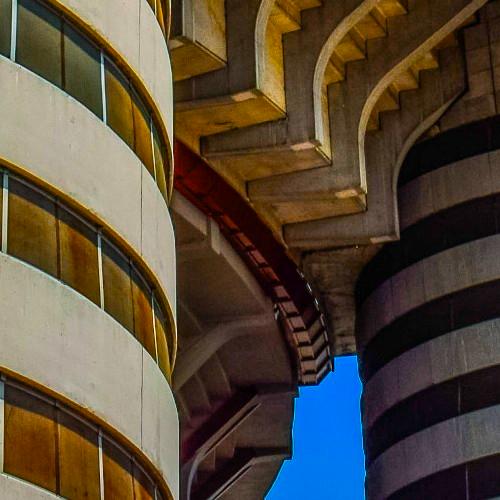} & \includegraphics[width=\gridimagewidth,valign=m]{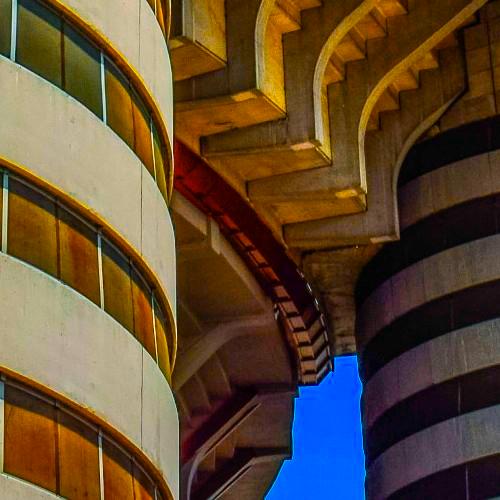} & \includegraphics[width=\gridimagewidth,valign=m]{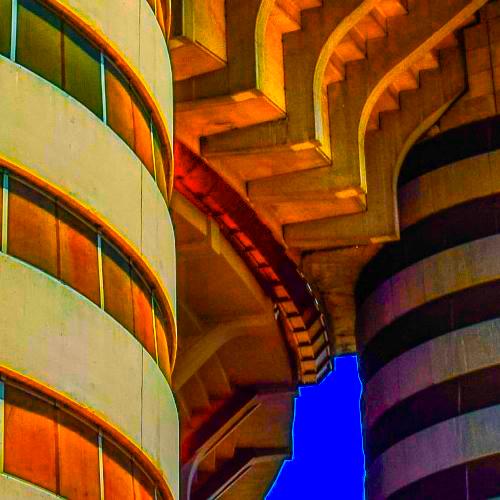} & \includegraphics[width=\gridimagewidth,valign=m]{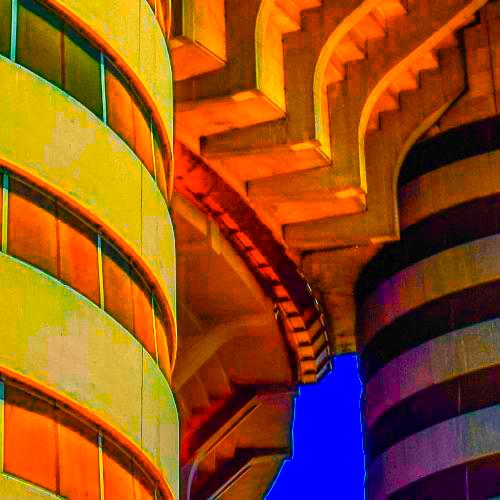} \\
\end{tabular}
}
\caption{Visualization of the degradation types belonging to the \textit{Color distortions} group for increasing levels of intensity.}
\label{fig:color_distortions}
\end{figure*}

\begin{figure*}
    \centering
    \setlength{\tabcolsep}{1pt}
    \Large
    \resizebox{\textwidth}{!}{ 
\begin{tabular}{C{6em}ccccc}
         & Level 1 & Level 2 & Level 3 & Level 4 & Level 5 \\
 JPEG2000 & \includegraphics[width=\gridimagewidth,valign=m]{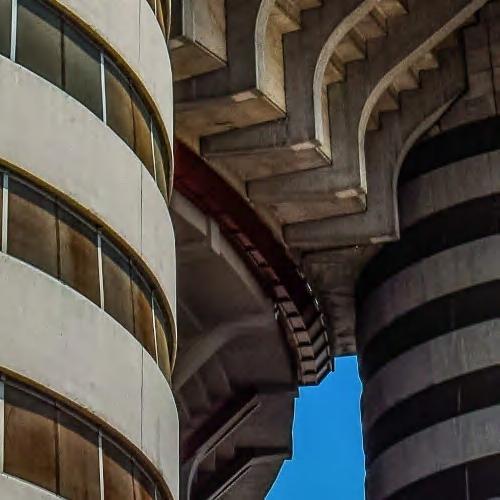} & \includegraphics[width=\gridimagewidth,valign=m]{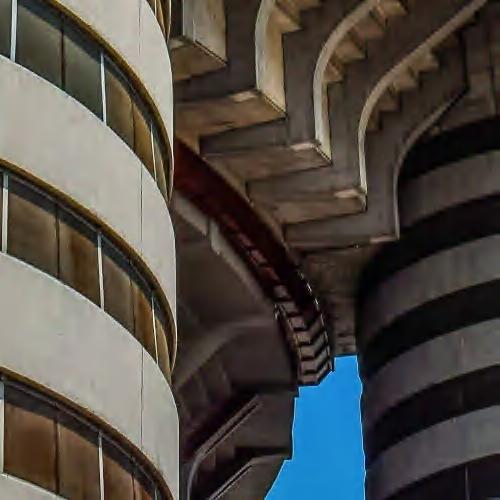} & \includegraphics[width=\gridimagewidth,valign=m]{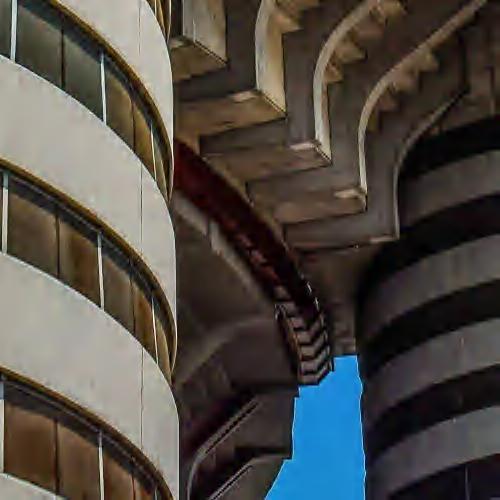} & \includegraphics[width=\gridimagewidth,valign=m]{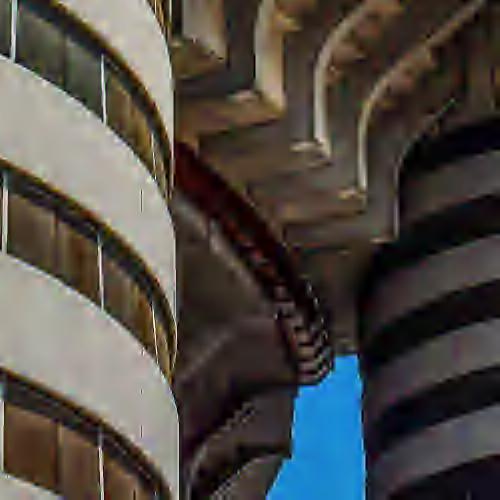} & \includegraphics[width=\gridimagewidth,valign=m]{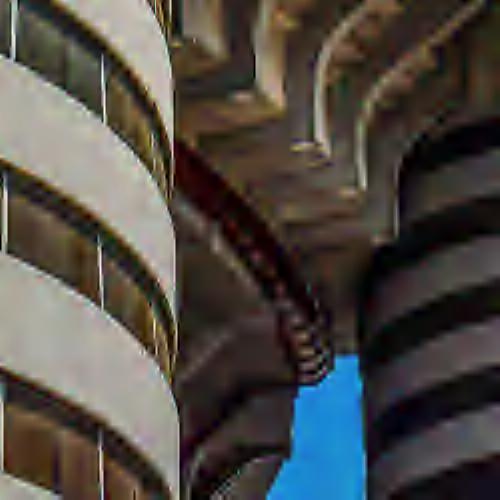} \\ [6.15ex]
 JPEG & \includegraphics[width=\gridimagewidth,valign=m]{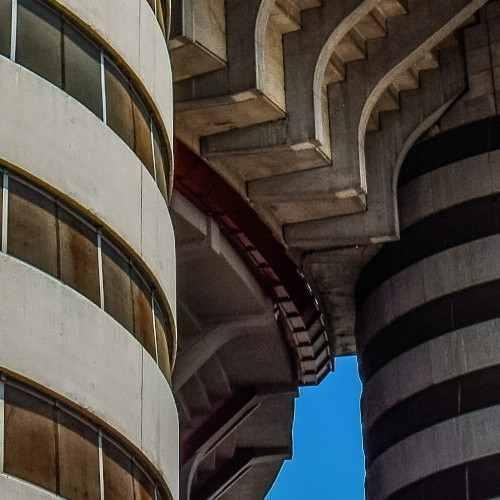} & \includegraphics[width=\gridimagewidth,valign=m]{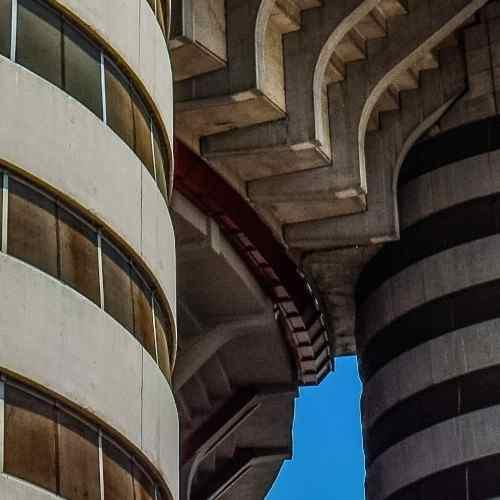} & \includegraphics[width=\gridimagewidth,valign=m]{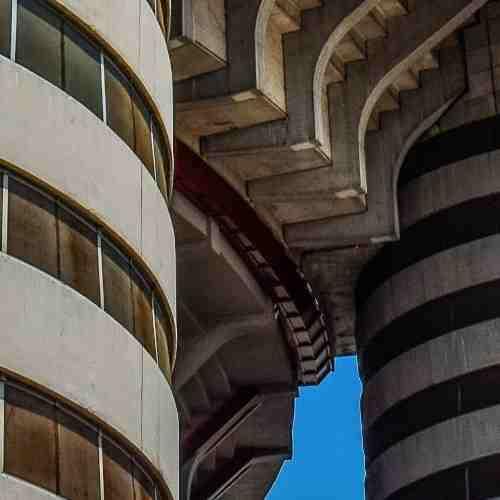} & \includegraphics[width=\gridimagewidth,valign=m]{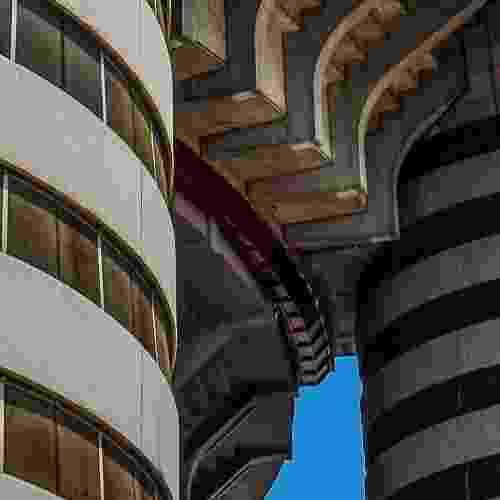} & \includegraphics[width=\gridimagewidth,valign=m]{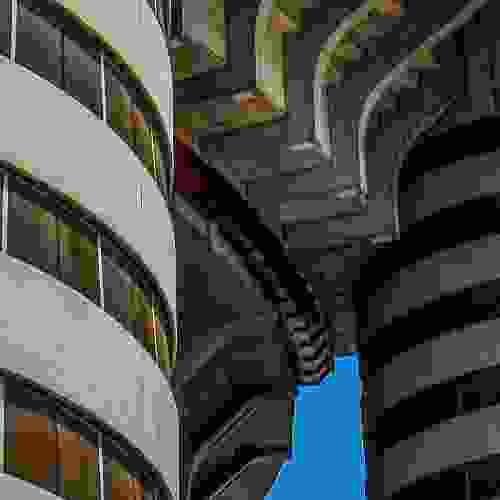} \\
\end{tabular}
}
\caption{Visualization of the degradation types belonging to the \textit{Compression} group for increasing levels of intensity.}
\label{fig:compression}
\end{figure*}

\begin{figure*}
    \centering
    \setlength{\tabcolsep}{1pt}
    \Large
    \resizebox{\textwidth}{!}{ 
\begin{tabular}{C{6em}ccccc}
         & Level 1 & Level 2 & Level 3 & Level 4 & Level 5 \\
 High sharpen & \includegraphics[width=\gridimagewidth,valign=m]{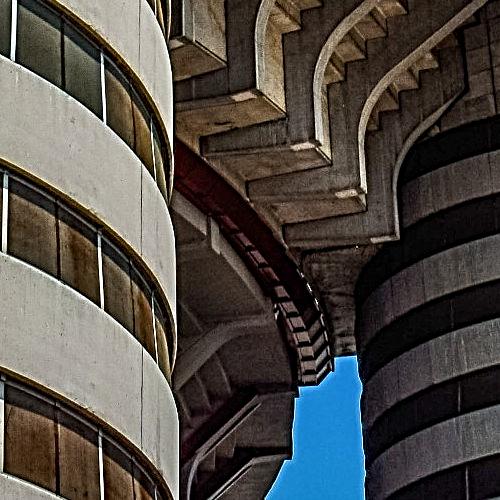} & \includegraphics[width=\gridimagewidth,valign=m]{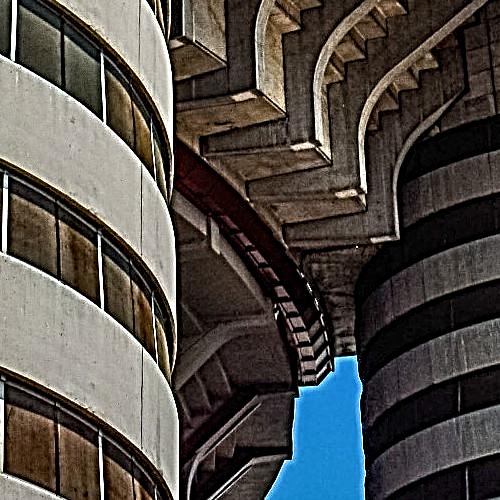} & \includegraphics[width=\gridimagewidth,valign=m]{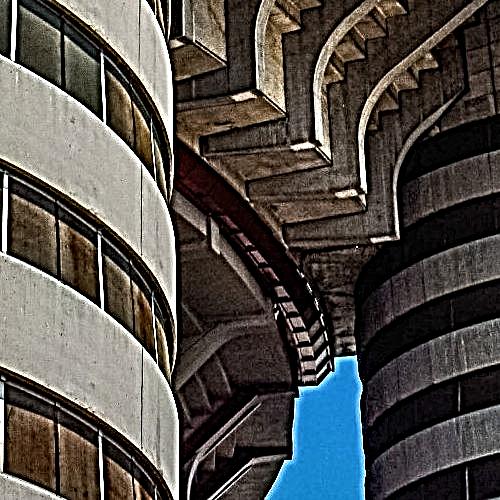} & \includegraphics[width=\gridimagewidth,valign=m]{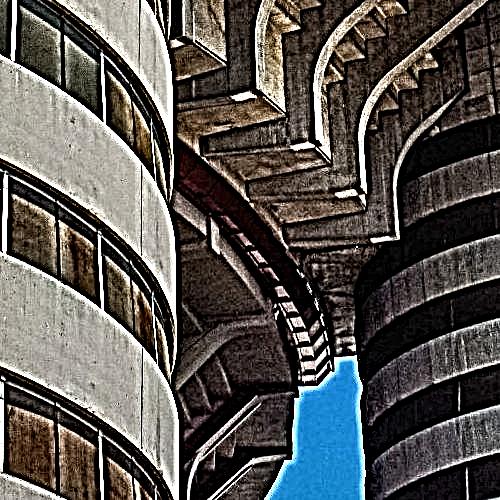} & \includegraphics[width=\gridimagewidth,valign=m]{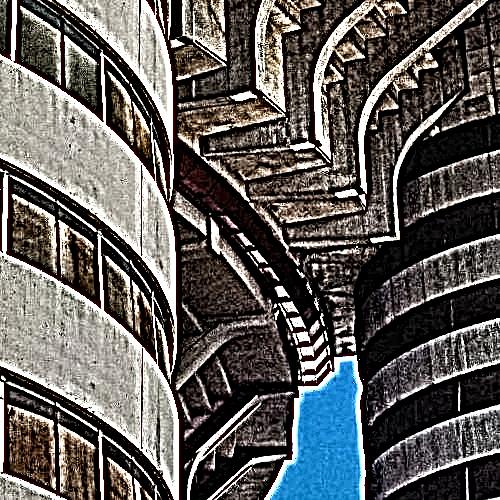} \\ [6.15ex]
 Nonlinear contrast change & \includegraphics[width=\gridimagewidth,valign=m]{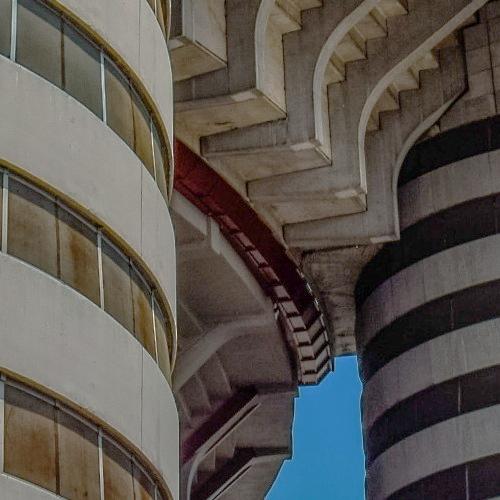} & \includegraphics[width=\gridimagewidth,valign=m]{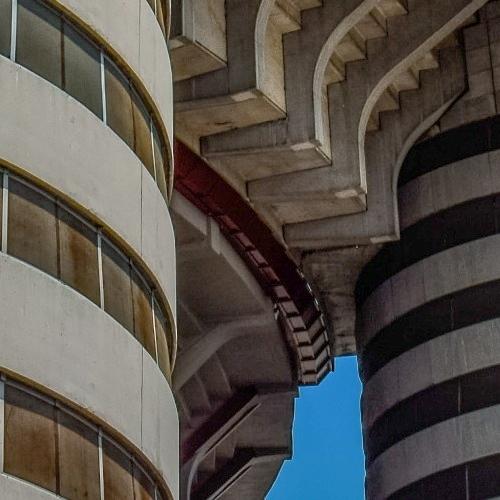} & \includegraphics[width=\gridimagewidth,valign=m]{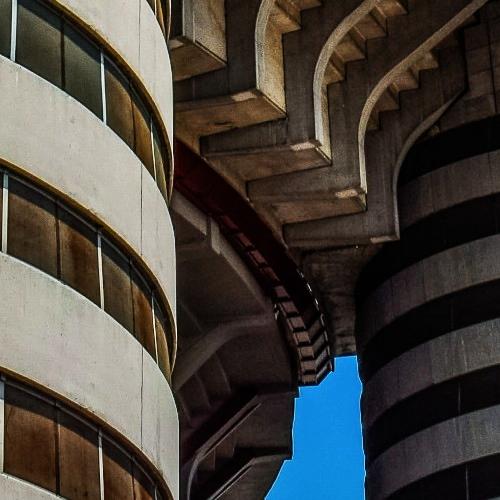} & \includegraphics[width=\gridimagewidth,valign=m]{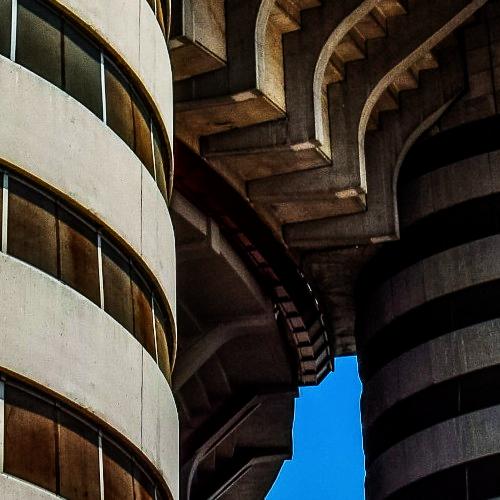} & \includegraphics[width=\gridimagewidth,valign=m]{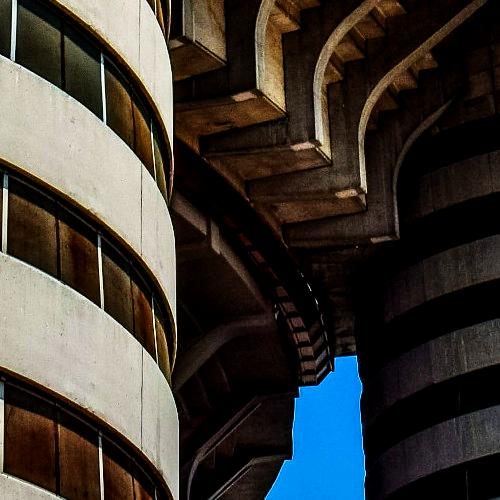} \\ [6.15ex]
 Linear contrast change & \includegraphics[width=\gridimagewidth,valign=m]{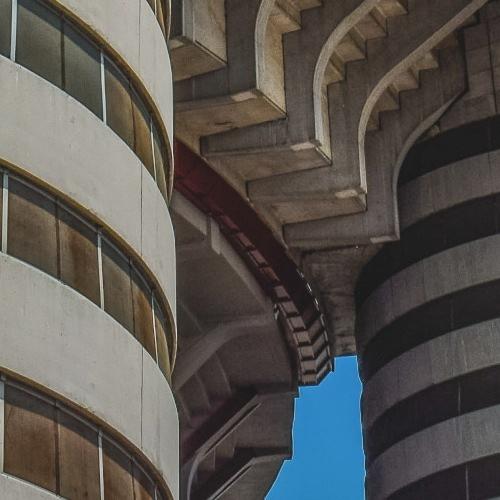} & \includegraphics[width=\gridimagewidth,valign=m]{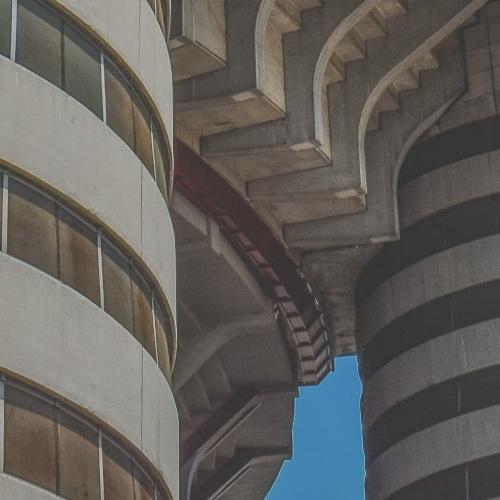} & \includegraphics[width=\gridimagewidth,valign=m]{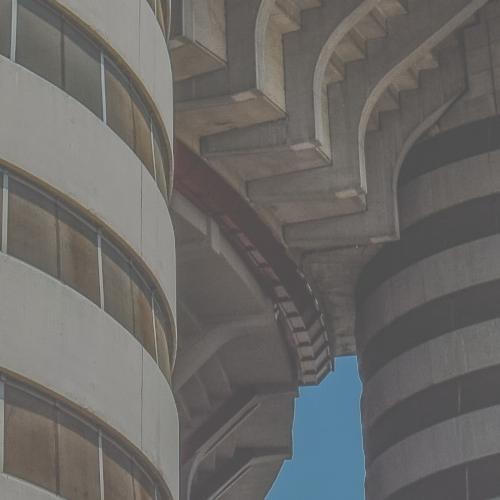} & \includegraphics[width=\gridimagewidth,valign=m]{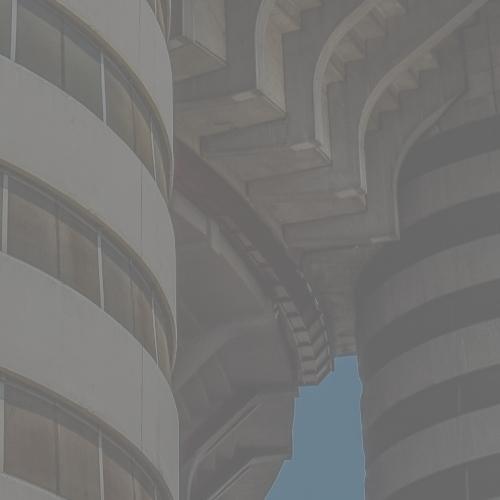} & \includegraphics[width=\gridimagewidth,valign=m]{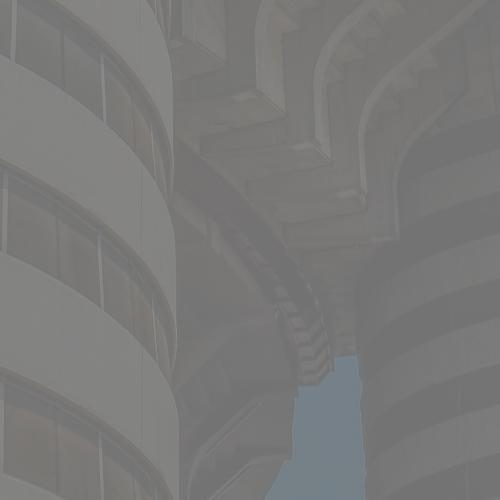} \\
\end{tabular}
}
\caption{Visualization of the degradation types belonging to the \textit{Sharpness \& contrast} group for increasing levels of intensity.}
\label{fig:sharpness_contrast}
\end{figure*}

Our image degradation model considers 24 different degradation types divided into the 7 distortion groups defined by the \kadid dataset \cite{kadid10k}. Each distortion has 5 levels of increasing intensity. \Cref{fig:brightness_change,fig:blur,fig:spatial_distortions,fig:noise,fig:color_distortions,fig:compression,fig:sharpness_contrast} shows the different levels of intensity for the degradations of each distortion group. The distortion types that we consider are mainly inspired by those of the \kadid dataset and are described in the list below:
\begin{enumerate}
    \item Brightness change:
        \begin{itemize}
            \item \textit{Brighten}: applies a sequence of color space transformations, curve adjustments, and blending operations to enhance the brightness of an input image, resulting in an output image with increased visual intensity;
            \item \textit{Darken}: similar to brighten operation, but it leads to a decreased visual intensity;
            \item \textit{Mean shift}: changes the average intensity of image pixels by adding a fixed amount to all the pixel values. Then, limits the resulting values to remain within the initial image range;
        \end{itemize}
    \item Blur:
        \begin{itemize}
            \item \textit{Gaussian blur}: filters every pixel of the image with a simple Gaussian kernel.
            \item \textit{Lens blur}: filters every pixel of the image with a circular kernel;
            \item \textit{Motion blur}: filters every pixel of the image with a linear motion blur kernel to simulate the effect of a moving camera or a moving object in the scene. Consequently, the image appears blurred in the direction of the motion;
        \end{itemize}
    \item Spatial distortions:
        \begin{itemize}
            \item \textit{Jitter}: randomly disperses image data by warping each pixel with small offsets;
            \item \textit{Non-eccentricity patch}: randomly extracts patches from the image and inserts them in random neighboring positions;
            \item \textit{Pixelate}: combines operations of downscaling and upscaling using nearest-neighbor interpolation;
            \item \textit{Quantization}: quantizes the image into $N$ uniform levels. The thresholds are computed dynamically using Multi-Otsu’s method \cite{liao2001fast};
            \item \textit{Color block}: randomly overlays homogeneous colored squared patches onto the image;
        \end{itemize}
    \item Noise:
        \begin{itemize}
            \item \textit{White noise}: adds Gaussian white noise to the image;
            \item \textit{White noise in color component}: converts the image to the YCbCr color space, then adds Gaussian white noise to each channel;
            \item \textit{Impulse noise}: adds salt and pepper noise to the image;
            \item \textit{Multiplicative noise}: adds speckle noise to the image;
        \end{itemize}
    \item Color distortions:
        \begin{itemize}
            \item \textit{Color diffusion}: converts the image to the LAB-color space, then applies Gaussian blur to each channel;
            \item \textit{Color shift}: randomly shifts the green channel and then blends it into the original image, masked by the normalized gradient magnitude of the original image;
            \item \textit{Color saturation 1}: converts the image to the HSV-color space and then multiplies the saturation channel by a factor;
            \item \textit{Color saturation 2}: converts the image to the LAB-color space, then multiply each color channel by a factor;
        \end{itemize}
    \item Compression:
        \begin{itemize}
            \item \textit{JPEG2000}: applies standard JPEG2000 compression to the image;
            \item \textit{JPEG}: applies standard JPEG compression to the image;
        \end{itemize}
    \item Sharpness \& contrast:
        \begin{itemize}
            \item \textit{High sharpen}: sharpens the image in the LAB-color space using unsharp masking;
            \item \textit{Nonlinear contrast change}: calculates a nonlinear tone mapping operation to manipulate the contrast of the image;
            \item \textit{Linear contrast change}: calculates a linear tone mapping operation to manipulate the contrast of the image;
        \end{itemize}
\end{enumerate}

\end{document}